\documentclass{nle}
\usepackage{booktabs}
\usepackage{tipa}
\usepackage{apalike}
\let\bibhang\relax
\usepackage{natbib}
\usepackage[algo2e,ruled,linesnumbered]{algorithm2e} 

\usepackage{tipa}
\usepackage{amsmath}
\usepackage{xcolor}
\usepackage{mathtools}
\usepackage{url}
\usepackage[caption=false]{subfig}
\usepackage{pifont}
\usepackage{lineno}
\usepackage{arabtex}
\usepackage{multicol}
\usepackage{multirow}
\usepackage{graphicx}
\usepackage{comment}
\usepackage{lineno}
\modulolinenumbers[5]
\newcommand{\cmark}{\ding{51}}%
\newcommand{\xmark}{\ding{55}}%

\newcommand{\squeezeup}{\vspace{-2.5mm}}
\title[A Clustering Framework for Lexical Normalization of Roman Urdu]
      {A Clustering Framework for Lexical Normalization of Roman Urdu}
        \author[A. Rafae and others]
   {A\ls B\ls D\ls U\ls L \ns R\ls A\ls F\ls A\ls E \ns K\ls H\ls A\ls N\\
   Stevens Institute of Technology, Hoboken, NJ, 07030, USA\\
   The Graduate Center Computer Science Department, City University of New York, 365 5th Ave, New York,\\ NY, 10016, USA
   \and
   A\ls S\ls I\ls M \ns K\ls A\ls R\ls I\ls M\\
   Lahore University of Management Sciences, D.H.A, Lahore Cantt., 54792, Lahore, Pakistan
   \and
   H\ls A\ls S\ls S\ls A\ls N \ns S\ls A\ls J\ls J\ls A\ls D\\
   Qatar Computing Research Institute, Hamad Bin Khalifa University, Doha, Qatar
   \and
   F\ls A\ls I\ls S\ls A\ls L \ns K\ls A\ls M\ls I\ls R\ls A\ls N\\
   Information Technology University, Arfa Software Technology Park, Ferozepur Road, Lahore,\\ Pakistan
   \and
   and\ns J\ls I\ls A \ns X\ls U\\
   Stevens Institute of Technology, Hoboken, NJ, 07030, USA\\
   The Graduate Center Computer Science Department, City University of New York, 365 5th Ave, New York,\\ NY, 10016, USA}

\pubyear{1998}

\begin{document}

\label{firstpage}
\maketitle

\begin{abstract}
Roman Urdu is an informal form of the Urdu language written in Roman script, which is widely used in South Asia for online textual content. It lacks standard spelling and hence poses several normalization challenges during automatic language processing. In this article, we present a feature-based clustering framework for the lexical normalization of Roman Urdu corpora, which includes a phonetic algorithm \emph{UrduPhone}, a string matching component, a feature-based similarity function, and a clustering algorithm \emph{Lex-Var}. UrduPhone encodes Roman Urdu strings to their pronunciation-based representations. The string matching component handles character-level variations that occur when writing Urdu using Roman script.
The similarity function incorporates various phonetic-based, string-based, and contextual features of words. The Lex-Var algorithm is a variant of the k-medoids clustering algorithm that groups lexical variations of words. It contains a similarity threshold to balance the number of clusters and their maximum similarity. The framework allows feature learning and optimization in addition to the use of pre-defined features and weights. We evaluate our framework extensively on four real-world datasets and show an F-measure gain of up to 15 percent from baseline methods. We also demonstrate the superiority of UrduPhone and Lex-Var in comparison to respective alternate algorithms in our clustering framework for the lexical normalization of Roman Urdu.
\end{abstract}
 
\section{Introduction}

Urdu, the national language of Pakistan, and Hindi, the national language of India,
jointly rank as the fourth most widely spoken language in the world \citep{ethnologue}.
Urdu and Hindi are closely related in morphology and phonology, but use different scripts: Urdu is written in Perso-Arabic script and Hindi is written in Devanagari script. Interestingly, for social media and short messaging service (SMS) texts, a large number of Urdu and Hindi speakers use an informal form of these languages written in Roman script, \emph{Roman Urdu}.

\seturdu

Since Roman Urdu does not have standardized spellings and is mostly used in informal communication, there exist many spelling variations for a word. For example, the Urdu word \<zndgy>\ [life] is written as \emph{zindagi}, \emph{zindagee}, \emph{zindagy}, \emph{zaindagee} and \emph{zndagi}. \setnone
The lack of standard spellings 
inflates 
the vocabulary of the language and causes sparsity problems. This results in poor performance of natural language processing (NLP) and text mining tasks, such as word segmentation \citep{durrani-hussain:2010:NAACLHLT}, part of speech tagging \citep{SajjadSchmid:09}, spell checking \citep{journals/lre/NaseemH07}, machine translation \citep{durrani-EtAl:2010:ACL}, and sentiment analysis \citep{paltoglou2012twitter}. For example, neural machine translation models are generally trained on a limited vocabulary. Non-standard spellings would result in a large number of words unknown to the model, which would result in poor translation quality.

Our goal is to perform \emph{lexical normalization}, which maps all spelling variations of a word to a unique form that corresponds to a single lexical entry.
This reduces data sparseness and improves the performance of NLP and text mining applications. 

\begin{figure}[t]
\centering
  \includegraphics[scale=0.25]{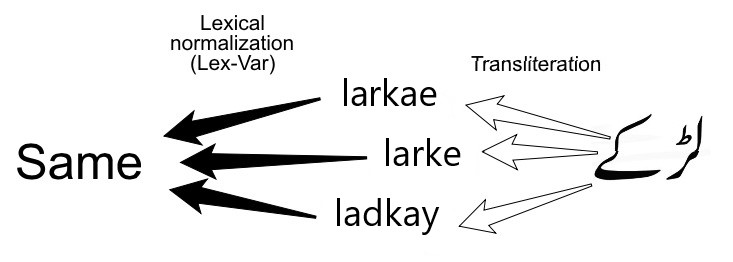}
   \caption{The lexicon can be varied due to informal writing, non-unified definition of transliteration, phonetic mapping etc.}
    \label{fig:Lex_var}
\end{figure}

One challenge of Roman Urdu normalization is 
lexical variations, which emerge through a variety of reasons such as
informal writing, inconsistent phonetic mapping, and non-unified transliteration. Compared to the lexical normalization of languages with a similar script like English, the problem is more complex than writing a language informally in the original script. For example, in English, the word \emph{thanks} can be written colloquially as \emph{thanx} or \emph{thx}, where the shortening of words and sounds into fewer characters is done in the same script. During Urdu to Roman Urdu conversion, two processes happen at the same time. (1) Various Urdu characters phonetically map to one or more Latin characters. (2)  The Perso-Arabic script is transliterated to Roman script. Since transliteration is a non-deterministic process, it also introduces spelling variations. Fig. \ref{fig:Lex_var} shows an example of an Urdu word \<l,rkE>\ [boys] that can be transliterated into Roman Urdu in three different ways (\emph{larke}, \emph{ladkay}, or \emph{larkae}) depending on the user's preference. Lexical normalization of Roman Urdu aims to map transliteration variations of a word to one standard form.

\setnone

\setnone

Another challenge is that Roman Urdu lacks a standard lexicon or labeled corpus for text normalization to use. Lexical normalization has been addressed for standardized or resource-rich languages like English, e.g.,  \citep{jin:2015:WNUT,han2013lexical,gouws2011unsupervised}. For such languages, the correct or the standard spelling of words is known, given the standard existence of the lexicon.
Therefore, lexical normalization typically involves finding the best lexical entry for a given word that does not exist in the standard lexicon. Thus, the proposed approaches aim to find the best set of standard words for a given non-standard word. On the other hand, 
Roman Urdu is an under-resourced language that does not have a standard lexicon. Therefore, it is not possible to distinguish between an in-lexicon and an out-of-lexicon word, and each word can potentially be a lexical variation of another. Lexical normalization of such languages is computationally more challenging than that of resource-rich languages.

Since we do not have a standard lexicon or labeled corpus for Roman Urdu lexical normalization, we cannot apply a supervised method. Therefore, we introduce an unsupervised clustering framework to capture lexical variations of words. In contrast to the English text normalization by \cite{sridhar2015unsupervised,Sproat17}, our approach does not require prior knowledge on the number of lexical groups or group labels (standard spellings). Our method significantly outperforms the state-of-the-art Roman Urdu lexical normalization using rule-based transliteration \citep{ahmed2009roman}.

In this work, we give a detailed presentation of our framework \citep{rafaeQUKSK15} with additional evaluation datasets, extended experimental evaluation, and analysis of errors. 
We develop an unsupervised feature-based clustering algorithm, \emph{Lex-Var}, that discovers groups of words that are lexical variations of one another. Lex-Var ensures that each word has at least a specified minimum similarity with the cluster's centroidal word.
Our proposed framework incorporates phonetic, string, and contextual features of words in a similarity function that quantifies the relatedness among words. We develop knowledge-based and machine-learned features for this purpose. The knowledge-based features include UrduPhone for phonetic encoding, an edit distance variant for string similarity, and a sequence-based matching algorithm for contextual similarity. We also evaluate various learning strategies for string and contextual similarities such as weighted edit distance and word embeddings.
%
For phonetic information, we develop \emph{UrduPhone}, an encoding scheme for Roman Urdu derived from Soundex. Compared to other available techniques that are limited to English sounds, UrduPhone is tailored for Roman Urdu pronunciations. For string-based similarity features, we define a function based on a combination of the longest common subsequence and edit distance metric.
For contextual information, we consider top-k frequently occurring previous and next words or word groups.
%
Finally, we evaluate our framework extensively on four Roman Urdu datasets: two group-chat SMS datasets, one Web blog dataset, and one service-feedback SMS dataset and measure performance against a manually developed database of Roman Urdu variations. Our framework gives an F-measure gain of up to 15\% as compared to baseline methods.

We make the following key \emph{contributions} in this paper: 
\squeezeup
\begin{itemize}
    \item We present a general framework for normalizing words in an under-resourced language that allows user-defined and machine-learned features for phonetic, string, and contextual similarity.
    \item We propose two different clustering frameworks including a k-medoids based clustering (Lex-Var) and an agglomerative clustering (Hierarchical Lex-Var)
    \item We present the first detailed study of Roman Urdu normalization.
    \item We introduce UrduPhone for the phonetic encoding of Roman Urdu words.
    \item We perform an error analysis of the results, highlighting the challenges of normalizing an under-resourced and non-standard language. 
    \item We have provided the source code for our lexical normalization framework.\footnote{\url{https://github.com/abdulrafae/normalization}}
\end{itemize}   

The remainder of this article is organized as follows. In Section \ref{sec:Task Definition}, we present our problem statement for the lexical normalization of an under-resourced language. In Section \ref{sec:Method}, we describe our clustering framework for the lexical normalization of an under-resourced language, including UrduPhone and Lex-Var. In Section \ref{sec:Experimental Evaluation}, we describe the evaluation criterion for the lexical normalization of Roman Urdu, describe the research experiments, and present the results and the error analysis. Section \ref{sec:previousWork} discusses the related work in the lexical normalization of informal language, and Section \ref{sec:conclusion} concludes the paper.

\section{Task Definition} \label{sec:Task Definition}
Roman Urdu is a transliterated form of the Urdu language written in Roman script. It does not have a standardized lexicon. That is, there is no standard spelling for words. Therefore, each word observed in a corpus can potentially be a variant of one or more of the other words appearing in the corpus. The goal of lexical normalization is to identify all spelling variations of a word in a given corpus. This challenging task involves normalizations associated with the following three issues:  (1) different spellings for a given word (e.g., \emph{kaun} and \emph{kon} for the word [who]); (2) identically spelled words that are lexically different (e.g., \emph{bahar} can be used for both [outside] and [spring]); and (3) spellings that match words in English (e.g., \emph{had} [limit] for the English word \emph{had}). The last issue arises because of code-switching between Roman Urdu and English, which is a common phenomenon in informal Urdu writing. 
People often write English phrases and sentences in Urdu conversations or switch language mid-sentence, e.g., \emph{Hi everyone. Kese ha aap log?} [Hi everyone. How are you people?]. 
In our work, we focus on finding common spelling variations of words (issue (1)), as this is the predominant issue in the lexical normalization of Roman Urdu and do not address issues (2) or (3) explicitly. 

\seturdu
Regarding issue (1), we note that while Urdu speakers generally transliterate Urdu script into Roman script, they also will often move away from the transliteration in favor of a phonetically closer alternative. A commonly observed example is the replacement of one or more vowels with another set of the vowels that has a similar pronunciation (e.g., \emph{janeaey} [to know] can also be written as \emph{janeey}). Here, the final characters 'aey' and 'ey' give the same pronunciation. Another variation of the previous word is \emph{janiey}. Now the character 'i' is replacing the character 'e'. In some cases, users will omit a vowel if it does not impact pronunciation, e.g., \emph{mehnga} [expensive] becomes \emph{mhnga} and similarly \emph{bohut} [very] becomes \emph{bht}. Another common example of this type of omission occurs with nasalized vowels. For example, the Roman Urdu word \emph{kuton} [dogs] is the transliteration of the Urdu word \<ktw.n>\ . But often, the final nasalized Urdu character \<.n>\ is omitted during conversion, and the Roman Urdu word becomes [kuto]. A similar case is found for words like \emph{larko} [boys], \emph{daikho} [see], \emph{nahi} [no] with final 'n' omitted. We incorporate some of these characteristics in our encoding scheme UrduPhone (See Section \ref{subsec:UrduPhone} and Table \ref{tab:soundex} for more details on UrduPhone, its rules, and for complete steps to generate encoding).
\setnone

We define the identification of lexical variations in an under-resourced language like Roman Urdu as follows: Given words $w_i$ ($i = 1, \ldots, N$) in a corpus, find the lexical groups $\ell_j$ ($j = 1, \ldots, K$) to which they belong. Each lexical group can contain one or more words corresponding to a single lexical entry and may represent different spelling variations of that entry in the corpus. In general, for a given corpus, the number of lexical groups $K$ is not known since no standardized lexicon is available. Therefore, we estimate it using normalization.


Clustering is expensive in the specific case of Roman Urdu normalization. Considering an efficient algorithm like k-means clustering, the computational complexity of lexical normalization is $O(NKT)$, where $T$ is the number of iterations required for clustering. 
By comparison, for languages like English with standardized lexicons, each out-of-vocabulary (OOV or not in the dictionary) word can be a variant of one or more in-vocabulary (IV) words. The computational complexity of lexical normalization in English (given by $O(K(N-K))$ where $K$ and $(N-K)$ are the numbers of IV and OOV words, respectively) is computationally less expensive than the lexical normalization of Roman Urdu.

\section{Method} \label{sec:Method}
In this section, we describe different components of our clustering framework. Section \ref{sec:lexC} formalizes our clustering framework including the algorithm developed. Section \ref{sec:sim_measure} defines a similarity function used in our clustering algorithm. In Section \ref{sec:features} we describe the features used in our system.

\subsection{Clustering Framework: Lex-Var}
\label{sec:lexC}

\begin{algorithm2e}[!h]
  \scriptsize
  \NoCaptionOfAlgo
\KwIn{$\mathcal{L}^\ast = \{\ell^\ast_1,\ell^\ast_2,\ldots ,\ell^{\ast}_{K^\ast}\}$ (initial clusters; see Table \ref{tab:exp}), $\mathcal{W} =\{w_1,w_2,\ldots ,w_N\}$ (words), $t$ (similarity threshold)}
\KwOut{$\mathcal{L} = \{\ell_1,\ell_2,\ldots ,\ell_{K}\}$ (predicted clusters) }
$\mathcal{L} = \mathcal{L}^\ast$\;
\Repeat{stop condition Satisfied}{
\tcc{Find cluster centroidal word}
    $\mathcal{C} = \emptyset$\;
    \For{$\forall~\ell_i ~ \in ~ \mathcal{L}$}{
        $\mathcal{R} = \emptyset$\;
        \For{$\forall ~ w_j ~ \in ~ \ell_i$}{  
            $r_j = 0$\;
            \For{$\forall\: w_k\: \in\: \ell_i$}{ 
                $r_j = r_j + S(w_j,w_k)$\;
            }
            $\mathcal{R} = \mathcal{R} \cup \{r_j\}$ \;
        }
        $m = \arg\max_j(r_j \in \mathcal{R})$\;
        $c_i = w_m$\;
        $\mathcal{C} = \mathcal{C} \cup \{c_i\}$\;
        $\ell_i = \emptyset$\;
    }
    \tcc{Assign word to clusters }
    \For{$\forall ~ w_i ~ \in ~ \mathcal{W}$}{ 
        $closest =$ null\;
        $maxSim = 0$\;
        \For{$\forall ~ c_j ~ \in ~ \mathcal{C}$}{
            \If{$S(w_i,c_j)> t$ \normalfont{and} $S(w_i,c_j)>maxSim$}{
                $maxSim = S(w_i,c_j)$\;
                $closest = c_j$\;
            }
        }
        \uIf(\tcp*[f]{Move word $w_i$ to cluster $\ell_{j}$}){$closest ~ != ~ $\normalfont{null}}{
             $\ell_j = \ell_j \cup \{w_i\} \mid closest \in \ell_j$\;}
    \Else(\tcp*[f]{Move word $w_i$ to new cluster $\ell_{\vert{\mathcal{L}}\vert+1}$}){
             $\ell_{\vert{\mathcal{L}}\vert+1} = \{w_i$\}\;
             $\mathcal{L} = \mathcal{L} \cup \{\ell_{\vert{\mathcal{L}}\vert+1}\}$\;
        }
    }
}
\caption{Algorithm \ref{alg:lexC}: Lex-Var}
\label{alg:lexC}
\end{algorithm2e}
We develop a new clustering algorithm, Lex-Var, for discovering lexical variations in informal texts. This algorithm is a modified version of the k-medoids algorithm \citep{dm_book} and incorporates an assignment similarity threshold, $t>0$, for controlling the number of clusters and their similarity spread. In particular, it ensures that all words in a group have a similarity greater than or equal to some threshold, $t$. It is important to note that the k-means algorithm cannot be used here because it requires that the means of numeric features describe the clustered objects. The standard k-medoids algorithm, on the other hand, uses the most centrally located object as a cluster's representative. 

Algorithm \ref{alg:lexC} gives the pseudo-code for Lex-Var. Lex-Var takes as input words ($\mathcal{W}$) and outputs lexical groups ($\mathcal{L}$) for these words. UrduPhone segmentation of the words gives the initial clusters. Lex-Var iterates over two steps until it achieves convergence. The first step finds the centroidal word $c_i$ for cluster $\ell_i$ as the word for which the sum of similarities of all other words in the cluster is maximal. In the second step, each non-centroidal word $w_i$ is assigned to cluster $\ell_j$ if $S(w_i, c_j)$ (see Section \ref{sec:sim_measure}) is maximal among all clusters and $S(w_i, c_j) > t$. If the latter condition is not satisfied (i.e., $S(w_i, c_j) \leq t$), then instead of assigning word $w_i$ to cluster $\ell_j$, it starts a new cluster. We repeat these two steps until a stop condition is satisfied (e.g., a fraction of words that change groups becomes less than a specified threshold). The computational complexity of Lex-Var is $O((n^2 + N)KT)$, where $n$ is the maximum number of words in a cluster, which is typically less than $N$.

\begin{figure}[!t]
    \centering
    \includegraphics[width=\textwidth]{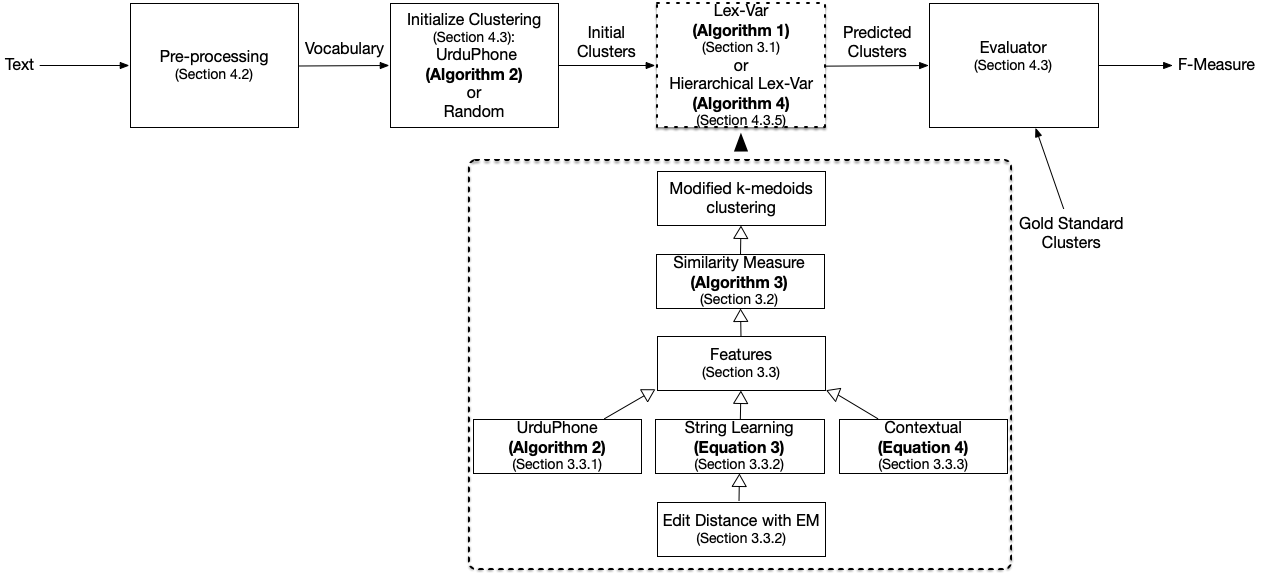}
    \caption{Flow Diagram for Lex-Var}
    \label{fig:Lex_var_flow}
\end{figure}

Fig. \ref{fig:Lex_var_flow} shows the details of our clustering framework. The first row of boxes shows the workflow of the system, and the area in the dotted square includes the modules used in our clustering method. The filled arrows indicate the outputs of the algorithms, and the unfilled arrows show modules that apply sub-modules.

After pre-processing the text, we normalize each word in the vocabulary. First, we initialize the clustering using random clustering or UrduPhone clusters. Then, based on the initial clusters, we apply (Hierarchical) Lex-Var algorithm to predict clusters. Finally, we compute the F- Measure based on the gold standard clusters to evaluate our prediction.

The Lex-Var algorithm applies a modified version of the k-medoids clustering, which uses a similarity measure that further consists of different features, including UrduPhone, String Learning, and Contextual feature. The edit distance is a sub-module of the string learning. We learn the substitution cost with various methods such as EM.

\subsection{Similarity Measure}
\label{sec:sim_measure}
We compute the similarity between two words $w_i$ and $w_j$ using the following similarity function:

\begin{equation}
\label{eq:similarity_function}
S(w_i, w_j) = \frac{ \sum_{f=1}^F \alpha^{(f)} \times \sigma_{ij}^{(f)}} {\sum_{f=1}^F \alpha^{(f)} }
\end{equation}
\noindent Here, $\sigma_{ij}^{(f)} \in [0, 1]$ is the similarity contribution made by feature $f$. $F$ is the total number of features. We will describe each feature  in Section~\ref{sec:features} in detail. $\alpha^{(f)} > 0$ is the weight of feature $f$. These weights are set to one by default and are automatically  optimized in Section \ref{sec:featureweights} and \ref{sec:weights}.  
The similarity function returns a value in the interval [0, 1] with higher values signifying greater similarity.

\subsection{Features}\label{sec:features}
The similarity function in Eq. \ref{eq:similarity_function} is instantiated with features representing each word. In this work, we use three features: phonetic, string, and contextual, which are computed based on rules or based on learning.

\begin{table}[!t]
    \caption{UrduPhone vs Soundex Encodings \label{tab:phonetic_comparison}}
    \scalebox{0.8}{
    \begin{minipage}{\textwidth}
    {\begin{tabular}{l | r | r}
        \toprule
        Word & Soundex Encoding & UrduPhone Encoding \\
        \midrule
        \emph{mustaqbil} [future] & M\_2\_3\_2 & M\_1\_2\_7\_9\_17\\
        \emph{mustaqil} [constant] & M\_2\_3\_2 & M\_1\_2\_7\_17\_0\\
        \midrule
        \emph{khirki} [window] & K\_6\_2\_0 & K\_19\_14\_7\_0\_0\\
        \emph{kursi} [chair] & K\_6\_2\_0 & K\_14\_1\_0\_0\_0\\ 
        \midrule
        \emph{ronak} [brightness] & R\_5\_2\_0 & R\_11\_7\_0\_0\_0\\
        \emph{rung} [color] & R\_5\_2\_0 & R\_11\_13\_0\_0\_0\\
        \midrule
		\emph{dimaagh} [brain] & D\_5\_2\_0 & D\_12\_13\_19\_0\_0\\
        \emph{dimaag} [brain] & D\_5\_2\_0 & D\_12\_13\_0\_0\_0\\
        \midrule
		\emph{please} & P\_4\_2\_0 & P\_17\_1\_0\_0\_0\\
        \emph{plx} & P\_4\_2\_0 & P\_17\_3\_0\_0\_0\\        
        \bottomrule
    \end{tabular}}
    \end{minipage}}
\end{table}

\begin{algorithm2e}[!t]
  \scriptsize
  \NoCaptionOfAlgo
\KwIn{$w=\{w_1,\cdots,w_n\}$, a word of length $n$}
\KwOut{$e=\{e_1,\cdots,e_6\}$, an encoding of length $6$}
$e[0]=uppercase(w[0])$;\\
$j=1$;\\
\For{$i=1\rightarrow n$}{
    \tcp{Discard duplicates}
    \If{$i+1\leq n \;\&\&\; w[i]==w[i+1]$}{
        continue;
    }
    \tcp{Discard Roman Urdu vowels (a,e,i,o,u,y)}
     \If{$w[i]==$`$a$'$\;||\; w[i]==$`$e$'$ \;||\; w[i]==$`$i$'$ \;||\; w[i]==$`$o$'$ \;||\; w[i]==$`$u$'$ \;||\; w[i]==$`$y$'}{
        continue;
    }
    \tcp{Encode character based on Table \ref{tab:soundex}}
    $e[j] = get\_encoding(w[i])$;\\
    $j++$;\\
}
\tcp{Add 0s if encoding length less than 6}
\While{$j\leq 6$}{
    $e[j] = 0$;\\
    $j++$;\\
}
\caption{Algorithm \ref{alg:uphone}: UrduPhone}
\label{alg:uphone}
\end{algorithm2e}

\seturdu
\begin{table}[!t]
    \caption{UrduPhone homophone mappings in Roman Urdu \label{tab:soundex}}\scalebox{.8}{
    \begin{minipage}{\textwidth}
    {\begin{tabular}{l | l|l|l}
      \toprule
      Characters & Urdu Alphabets & IPA\footnotemark & Example \\
      \midrule
      q,k & \RL{q} , \RL{k} & \textipa{[q]}, \textipa{[k]}  & \emph{qainchi} [scissors], \emph{kitab} [book] \\
      c,sh,s & \RL{s} , \RL{^s} , \RL{.s} , \RL{_t} & \textipa{[s]}, \textipa{[\textesh]}, \textipa{[s]}, \textipa{[s]} & \emph{shadi} [wedding], \emph{sadi} [simple]\\
      z,x & \RL{z} , \RL{_d} , \RL{.z} , \RL{.d} & \textipa{[z]}, \textipa{[z]}, \textipa{[z]}, \textipa{[z]} & \emph{zameen} [earth], \emph{xar} [gold]\\
      zh &  \RL{^z}  & \textipa{[Z]} & \emph{zhalabari} [hail]\\
      kh & \RL{_h} & \textipa{[x]} & \emph{zakhmi} [injured]\\
      d & \RL{d} , \RL{,d} & \textipa{[\|[{d}]}, \textipa{[\textrtaild]} & \emph{dahi} [yogurt], \emph{doob} [sink]\\
      t & \RL{t} , \RL{,t} , \RL{.t} & \textipa{[\|[{t}]}, \textipa{[\textrtailt]}, \textipa{[\|[{t}]} & \emph{tareef} [praise], \emph{timatar} [tomato]\\
      m & \RL{m} & \textipa{[m]} & \emph{maut}  [death]\\
      j & \RL{j} & \textipa{[\|c{dZ}]} & \emph{jism} [body]\\
      g & \RL{g}  & \textipa{[g]} & \emph{gol} [circular]\\
      f & \RL{f} & \textipa{[f]} & \emph{fauj} [army]\\
      b & \RL{b} & \textipa{[b]} & \emph{bjli} [lightening]\\
      p & \RL{p} & \textipa{[p]} & \emph{pyaz} [onion]\\
      l & \RL{l} & \textipa{[l]} & \emph{lafz} [word]\\
      ch & \RL{^c} & \textipa{[\|c{t\textesh}]} & \emph{chehra} [face]\\
      h & \RL{.h} , \RL{,h} , \RL{h} & \begin{minipage}{30mm} \textipa{[h, H ]}, \textipa{[h, H, \o]} , \\ \textipa{[\super{h}, \super{H}]} \end{minipage} & \begin{minipage}{30mm}\emph{haal} [present],\\ \emph{bahar} [spring], \emph{phal} [fruit]\end{minipage}\\
      n & \RL{n} , \RL{.n} & \textipa{[n, \textltailn, \textrtailn, \ng]}, \textipa{[ \~{} ]} & \emph{nazar} [sight], \emph{larkioun} [girls]\\
      r & \RL{r}, \RL{,r}  & \textipa{[r]}, \textipa{[\textrtailr]} & \emph{risala} [magazine], \emph{guriya} [doll]\\
      w,v &  \RL{w} , \RL{`} & \begin{minipage}{30mm} \textipa{[V, u:, o:, \textopeno:]}, \\ \textipa{[a:, o:, e:, P, Q, \o]} \end{minipage} & \emph{waqt} [time], \emph{vada} [promise]\\
      bh &  \RL{bh}  & \textipa{[b\super{h}]} & \emph{bhaag} [run]\\
      ph &  \RL{ph}  & \textipa{[p\super{h}]} & \emph{phool} [flower]\\
      jh &  \RL{jh}  & \textipa{[\|c{dZ}\super{h}]} & \emph{bojh} [weight], \emph{boj} [weight]\\
      th &  \RL{th} , \RL{,th}   & \textipa{[\|[t\super{h}]}, \textipa{\![t\super{h}]} & \emph{thapki} [pat], \emph{thokar} [stumble]\\
      dh &  \RL{dh} , \RL{,dh}  & \textipa{[\|[{d}]\super{h}]}, \textipa{[\textrtaild\super{h}]} & \emph{udhar} [loan], \emph{dhool} [drum]\\
      rh &  \RL{rh} , \RL{,rh}  & \textipa{[\:r\super{h}]}, \textipa{[\textrtailr\super{h}]} & \emph{rhnuma} [guide], \emph{barhna} [to grow] \\
      gh &  \RL{.g}  & \textipa{[G]} & \emph{ghalat} [wrong]\\
      a,i,e,o,u,y &  \RL{a} , \RL{y} , \RL{E} , \RL{w}, \RL{`}, \RL{'} & \begin{minipage}{30mm} \textipa{[a:, P, \o]}, \textipa{[j, i:, a:]}, \\ \textipa{[\textepsilon:,e:]}, \textipa{[V, u:, o:, \textopeno:]}, \\ \textipa{[a:, o:, e:, P, Q, \o]},\\ \textipa{[P, \o]} \end{minipage} & \begin{minipage}{35mm} \emph{aam} [mango],\\ \emph{ilm} [knowledge], \emph{ullu} [owl] \end{minipage}\\
      \bottomrule
    \end{tabular}}
    \end{minipage}}
\end{table}
\addtocounter{footnote}{0}
\footnotetext{\url{https://en.wikipedia.org/wiki/Urdu_alphabet}}
\setnone
\subsubsection{UrduPhone}
\label{subsec:UrduPhone}

We propose a new phonetic encoding scheme, \emph{UrduPhone}, tailored for Roman Urdu. Derived from Soundex \citep{soundexbook,DBLP:journals/csur/HallD80}, UrduPhone encodes consonants by using similar sounds in Urdu and English.  
UrduPhone differs from Soundex in two ways: 

1) UrduPhone's encoding of words contains six characters as opposed to four in Soundex. An increase in encoding length reduces the possibility of mapping semantically different words to one form. 
Soundex maps different words to a single encoding, which, due to the limited encoding length, can cause errors when trying to find correct lexical variations. See Table \ref{tab:phonetic_comparison} for some examples of the differences.
For example, \emph{mustaqbil} [future] and \emph{mustaqil} [constant] encode to one form, \emph{MSTQ}, in Soundex but to two different forms using UrduPhone encoding. In a limited number of cases, UrduPhone increases ambiguity by mapping lexical variations of the same word into different encodings, as in the case of \emph{please} and \emph{plx}. Since these words share a similar context, 
these variations will
map to one cluster with the addition of contextual information. This is also shown during our experiments.

2) We introduce homophone-based groups, which are mapped differently in Soundex. There are several Urdu characters, which map to the same Roman form. For example,
\emph{samar} [reward], \emph{sabar} [patience], and \emph{saib} [apple], all start with different Urdu characters that have an identical Roman representation: \emph{s}. 
We group together homophones 
such as \emph{w, v} as in \emph{ taweez, taveez } [amulet] and \emph{z, x} as in \emph{lolz, lolxx} [laughter] or \emph{zara, xara} [a bit].
One common characteristic with transliteration from Urdu to Roman script is the omission of the Roman character 'h'. For example, the same Urdu word maps to both the Roman words \emph{samajh} \& \emph{samaj} [to understand]. This is especially true in the case of digraphs representing Urdu aspirates such as \emph{dh, ph, th, rh, bh, jh, gh, zh, ch}, and \emph{kh}. A problem arises when the longest common subsequence in words (if 'h' is omitted) causes overlaps such as (\emph{khabar} [news], \emph{kabar} [grave]) and (\emph{gari} [car], \emph{ghari} [watch]).  
Also, when \emph{sh} comes at the end of a word, as in \emph{khawhish, khawhis} [wish]; when 'h' is omitted, the sound is mapped to the character \emph{s}. 
\seturdu Similarly, if there is a transcription error, such as \emph{dushman} [enemy] becomes \emph{dusman}, the UrduPhone encoding is identical. Here, the omission of 'h' causes an overlap of the characters \<s>\ and \<^s>\ . \setnone

The second column of Table \ref{tab:phonetic_comparison} shows a few examples of Soundex encodings of Roman Urdu words. In some cases, Soundex maps two semantically different words to one code, which is undesirable in the task of lexical normalization. 
Table \ref{tab:soundex} shows a complete list of homophone-based mapping introduced in UrduPhone, and Algorithm \ref{alg:uphone} shows the process to encode a word into an UrduPhone encoding. Then, we compute the phonetic similarity of words $w_i$ and $w_j$ using Eq. \ref{eq:phonetic}.

\begin{align}
    \sigma_{ij}^{P} = 
    \begin{cases}
    1 & \text{if } UrduPhone(w_i)==UrduPhone(w_j) \\
        0 & \text{otherwise}
    \end{cases}
    \label{eq:phonetic}
\end{align}

\subsubsection{Learning String-similarity}
\label{sec:string}
The lexical variations of a word may have a number of overlapping sub-word units, e.g., spelling variations of \emph{zindagi} [life] include \emph{zindagee}, \emph{zindagy}, \emph{zaindagee} and \emph{zndagi} with many overlapping sub-word units. To benefit from this overlap, we define a string similarity function as follows:

\begin{equation}
\label{eq:string_similarity}
\sigma_{ij}^{S} = \frac{\mathit{lcs}(w_i, w_j) } {\min[\mathit{len}(w_i), \mathit{len}(w_j)] + \mathit{edist}(w_i, w_j)}
\end{equation}
\noindent Here, $\mathit{lcs}(w_i, w_j)$ is the length of the longest common subsequence in words $w_i$ and $w_j$,  $\mathit{len}(w_i)$ is the length of word $w_i$, and $\mathit{edist}(w_i, w_j)$ is the edit distance between words $w_i$ and $w_j$. 
\noindent\paragraph{Edit Distance:}
The edit distance allows insertion, deletion and substitution operations. We obtain the cost of edit distance operations in two ways:

\paragraph{\textbf{Manually Defined}} -- In a naive approach, we consider the cost of every operation to be equal and set them to $1$. We refer to this edit distance cost as edist$_{man}$. This technique has a downside of considering all operations equally necessary, which is an erroneous assumption. For example, the substitution cost of a Roman character 'a' to 'e' should be less than the cost of 'a' to 'z' because both 'a' and 'e' have related sounds in some contexts. It is possible to use these characters alternatively when transliterating from Perso-Arabic script to Roman Script. 

\paragraph{\textbf{Automatically Learning Edit Distance Cost}} -- In this approach, we automatically learn the edit distance cost from the data. Consider a list of word pairs where one word is a lexical variation of another word. One can automatically learn the character alignments between them using an EM algorithm. The inverse character alignment probability serves as the cost for the edit distance operations. 

In our case, we do not have a cleaned list of word pairs to learn character alignments automatically. Instead, we try to learn these character alignments from the noisy training data. To do this, we build a list of candidate word pairs by aligning every word to every other word in the corpus as a possible lexical variation. We split the words into characters and run the word-aligner GIZA++ \citep{och2003systematic}. Here, the word-aligner considers every character as a word and every word as a sentence. We use the learned character alignments with one minus their probability as the cost for the edit distance function. We refer to this edit distance cost as edist$_{giza}$.   

Since the model learns the cost from the noisy data, likely, it is not a good representative of the accurate edit distance cost that would be learned from the cleaned data. In our alternative method, we automatically refine the list of candidate pairs and learn character alignments from it. In this approach, we consider the problem of lexical variations as a \textbf{transliteration mining} problem \citep{sajjad:acl11}, where, given a list of candidate word pairs, the algorithm automatically extracts word pairs that are transliterations of each other. For this purpose, we use the unsupervised transliteration mining model of \cite{sajjad2017statistical}, who define the model\footnote{\url{https://github.com/hsajjad/transliteration_mining}} as a mixture of a transliteration sub-model and a non-transliteration sub-model. The transliteration sub-model generates the source and target character sequences jointly and can model the dependencies between them. The non-transliteration model consists of two monolingual character sequence models that generate source and target strings independently of each other. The parameters of the transliteration sub-model are uniformly initialized and then learned during EM training of the complete interpolated model. 
During the training process, the model penalizes character alignments that are less likely to be part of a transliteration pair and favors character alignments that are likely to be part of a transliteration pair.

We train the unsupervised transliteration miner on our candidate list of word pairs, similar to the GIZA++ training. Then, we learn the character alignments. We then use these character alignments with one minus their probability as the cost for the edit distance metric. We refer to this cost as edist$_{miner}$.

\subsubsection{Context Information} \label{sec:context_feature}

We observe that non-standard variants of a standard word have similar contexts.
For example, \emph{truck} and \emph{truk} will be used in similar contexts, which might be very different from \emph{cat}. We used this idea to define a contextual similarity measure between two words. We compare the top-k frequently occurring preceding (previous) and following (next) words' features of the two words in the corpus. The previous and next word's features can be each word's ID, UrduPhone ID, or cluster/group ID (based on initial clustering of the words). 

Let $a_1^i, a_2^i, \ldots, a_5^i$ and $a_1^j, a_2^j, \ldots, a_5^j$ be the features (word IDs, UrduPhone IDs, or cluster IDs) for the top-5 frequently occurring words preceding word $w_i$ and $w_j$, respectively. We use the similarity between the two words based on this context as defined by \cite{rank_weight}:

\begin{equation}
\label{eq:rank_corr}
\sigma_{ij}^{C} = \frac{\sum_{k=1}^5 \rho_k} {\sum_{k=1}^5 k}
\end{equation}
\noindent Here, $\rho_k$ is zero for any $a_k^i$ (i.e., the $k$th word in the context of $w_i$) when there exists no match in $a_*^j$ (i.e., in the context of word $w_j$). Otherwise, $\rho_k = 5 - \max[k, l] -1$ where $a_k^i = a_l^j$ and $l$ is the highest rank (smallest integer) at which a previous match has not occurred. In other words, this measure is the normalized sum of rank-based weights for matches in the two sequences, with more importance given to those occurring in higher ranks. Note that contextual similarity can be computed even if the context sizes of the two words are different, an essential step as a word may not have 5 distinct words preceding it in the corpus.

\begin{algorithm2e}[!t]
  \scriptsize
  \NoCaptionOfAlgo
\KwIn{$w_i,w_j$ (input words), $F$ (set of features used), $\alpha$ (set of feature weights)}
\KwOut{$sim$ (similarity between $w_i$ \& $w_j$)}
$total\_weight=0$;\\
$sum = 0$;\\
\If{$phonetic\in F$}{
    $encoded_i=urduphone(w_i)$\;
    $encoded_j=urduphone(w_j)$\;
    $\sigma_{ij}^{P}= phonetic\_sim(encoded_i==encoded_j)$; \tcp{see Eq. \ref{eq:phonetic}}
    $sum = sum + \alpha^{P}\times \sigma_{ij}^{P}$\;
    $total\_weight=total\_weight+\alpha^{P}$\;
}
\If{$string \in F$}{
    $\sigma_{ij}^{S} = string\_sim(w_i,w_j)$; \tcp{see Eq. \ref{eq:string_similarity}}
    $sum = sum + \alpha^{S}\times \sigma_{ij}^{S}$\;
    $total\_weight=total\_weight+\alpha^{S}$\;
}
\If{$context \in F$}{
    $prev_i=top5prev(w_i)$\;
    $prev_j=top5prev(w_j)$\;
    $\sigma_{ij}^{C_1}= context\_sim(prev_i,prev_j)$; \tcp{see Eq. \ref{eq:rank_corr}}
    $sum = sum + \alpha^{C_1}\times \sigma_{ij}^{C_1}$\;
    $total\_weight=total\_weight+\alpha^{C_1}$\;
    $next_i=top5next(w_i)$\;
    $next_j=top5next(w_j)$\;
    $\sigma_{ij}^{C_2}= context\_sim(next_i,next_j)$; \tcp{see Eq. \ref{eq:rank_corr}}
    $sum = sum + \alpha^{C_2}\times \sigma_{ij}^{C_2}$\;
    $total\_weight=total\_weight+\alpha^{C_2}$\;
}
\If{$word2vec \in F$}{
    $vec_i = word\_vector(w_i)$; \tcp{see Section \ref{sec:lexc_vs_others}}
    $vec_j = word\_vector(w_j)$\;
    $\sigma_{ij}^{W}=cosine(vec_i,vec_j)$\;
    $sum = sum + \alpha^{W}\times \sigma_{ij}^{W}$\;
    $total\_weight=total\_weight+\alpha^{W}$\;
}
\If{$2skip1gram \in F$}{
    $sigma_{ij}^{G} = 2skip1gram(w_i,w_j)$; \tcp{see Algorithm \ref{alg:skipgram}}
    $sum = sum + \alpha^{G}\times \sigma_{ij}^{G}$\;
    $total\_weight=total\_weight+\alpha^{G}$\;
}
$sim = \frac{sum}{total\_weight}$\;
\caption{Algorithm \ref{alg:sim_measure}: Similarity measure with weighted feature combination}
\label{alg:sim_measure}
\end{algorithm2e}

We combine all the features using our similarity measure from Eq. \ref{eq:similarity_function}. The code for combining a set of features is in Algorithm \ref{alg:sim_measure}.

\subsection{Parameter Optimization}\label{sec:featureweights}

\textbf{The feature weights} $\alpha^{(f)}$ used to measure word similarity in Eq.~\ref{eq:similarity_function} can be tuned to optimize prediction accuracy. For example, by changing the weights in our clustering framework (see Eq. \ref{eq:similarity_function}), we can make contextual similarity more prominent (by increasing the weight $\alpha^{C}$) so that words with the same UrduPhone encoding but different contexts are placed in separate clusters (see discussion in Section \ref{sec:error_analysis}). But, we also test with other weight combinations and features, including using both word IDs and UrduPhone IDs to represent the top-5 most frequently occurring previous and next words (rather than just one representation as used in other experiments). We identify corresponding weights for contexts based on word IDs and UrduPhone IDs as $\alpha^{C_1}$ and $\alpha^{C_2}$, respectively. The weights for the phonetic and the string features are $\alpha^{P}$ and $\alpha^{S}$, respectively. 

We also optimize $n$ variables to maximize an objective function using the Nelder-Mead method \citep{NelderMead65}. We use the Nelder-Mead method to maximize the F-measure by optimizing the feature weights of our Similarity function in Eq. \ref{eq:similarity_function}, as well as \textbf{the hyperparameter, threshold $t$,} in Line 21 of Algorithm \ref{alg:lexC}. We apply 10-fold cross-validation on the SMS (small) dataset (Table \ref{tab:weights_change_exp}). We will describe the results in Section~\ref{sec:weights}.

\section{Experiments} \label{sec:Experimental Evaluation}
 
In this section, we first describe our evaluation setup and the datasets used for the experiments. Later, we present the results.

\subsection{Evaluation Criteria}
\label{subsubsec:BCubedMeasure}
Since the lexical normalization of Roman Urdu is equivalent to a clustering task, we can adopt measures for evaluating clustering performance. We need a gold standard database defining the correct groupings of words for evaluation. This database contains groups of words such that all words in a given group are considered lexical variations of a lexical entry. In clustering terminology, words within a cluster are more similar than words across clusters. On the other hand, we typically use the accuracy (i.e., the proportion of OOV words that correctly match IV words) to evaluate the lexical normalization of a standardized language like English. This measure is appropriate because we know the IV words and can be compared to every OOV word. 

\citet{bCubed98} discussed measures for evaluating clustering performance and recommend the use of BCubed precision, recall, and F-measure. These measures possess all four desirable characteristics for clustering evaluation (homogeneity, completeness, rag bag, and cluster size vs. the number of clusters -- see \cite{M95-1005} for details). In the context of the lexical normalization of non-standard languages, they provide the additional benefit that they are computed for each word separately and then averaged for all words. For example, if a cluster contains all variants of a word and nothing else, then it is considered homogeneous and complete, and this is reflected in its performance measures. These measures are robust in the sense that incorporating small impurities in an otherwise pure cluster impacts the measures significantly (rag bag characteristic), and the trade-off between cluster size and the number of clusters is reflected appropriately. Other clustering evaluation measures do not possess all these characteristics and, in particular, commonly-used measures like entropy and purity are not based on individual words. 

Let $\mathcal{L} = \{\ell_{1},\cdots ,\ell_{K}\}$ be the set of output clusters and $\mathcal{L}^{\prime} = \{\ell_{1}^{\prime},\cdots ,\ell_{K}^\prime\}$ be the set of actual or correct clusters in the gold standard. We define correctness for word pair $w_i$ and $w_j$ as

\begin{equation}
C(w_i,w_j) =
\begin{cases}
1 & \text{iff } (w_i \in \ell_m, w_j \in \ell_m)\text{ and }(w_i \in \ell_n^\prime, w_j \in  \ell_n^\prime) \\
0 & \text{otherwise}
\end{cases}
\label{eq:correctness}
\end{equation}
\noindent In other words, $C(w_i, w_j)=1$ when words $w_i$ and $w_j$ appear in the same cluster ($\ell_m$) of the clustering and the same cluster ($\ell_n^\prime$) of the gold standard; otherwise, $C(w_i, w_j)=0$. By definition, $C(w_i, w_i)=1$. 

The following expressions give the BCubed precision $P(w_i)$ and recall $R(w_i)$ for a word $w_i$: 

\begin{equation}
P(w_i) =
\frac{\sum_{j=1}^{N} C(w_i,w_j) }{\left\vert{\ell_m}\right\vert}
\label{eq:precision}
\end{equation}
\begin{equation}
R(w_i) =
\frac{\sum_{j=1}^{N} C(w_i,w_j) }{\left\vert{\ell^\prime_m}\right\vert}
\label{eq:recall}
\end{equation}
\noindent Here, $\ell_m$ and $\ell_m^\prime$ identify the cluster in the clustering and gold standard, respectively, that contain word $w_i$. The summation for Eq. \ref{eq:precision} \& Eq. \ref{eq:recall} is over all the words $j$. Finally, we define the BCubed F-measure $F(w_i)$ of word $w_i$ in the usual manner as: 

\begin{equation}
F(w_i) = 2\times \frac{P(w_i)\times R(w_i)}{P(w_i)+R(w_i)}
\label{eq:fmeasure}
\end{equation}
\noindent We compute the overall BCubed precision, recall, and F-measure of the clustering as the average of the respective values for each word. For example, we calculate the F-measure of the clustering as $\frac{\sum_{i=1}^N F(w_i)}{N}$. 

\subsection{Datasets}
We utilize four datasets in our experimental evaluation. The first and second datasets, SMS (small) and SMS (large), are obtained from Chopaal, an internet-based group SMS service.\footnote{\url{http://chopaal.org}} These two versions are from two different time periods and do not overlap. The third dataset, Citizen Feedback Monitoring Program (CFMP) dataset, is a collection of SMS messages sent by citizens as feedback on the quality of government services (e.g., healthcare facilities, property registration).\footnote{\url{http://cfmp.punjab.gov.pk/}}  The fourth dataset, Web dataset, is scraped from Roman Urdu websites on news,\footnote{\url{http://www.shashca.com}, \url{http://stepforwardpak.com/}} poetry,\footnote{\url{https://hadi763.wordpress.com/}} SMS,\footnote{\url{http://www.replysms.com/}} and blogs.\footnote{\url{http://roman.urdu.co/}} Unless mentioned otherwise, the SMS (small) dataset is used for the experiment.
All four datasets are pre-processed with the following steps: (1) Remove single-word sentences; 
(2) add tags to URLs, email addresses, time, year, and numbers with at least four digits; (3) Collapse more than two repeating groups to only two (e.g., \emph{hahahaha} to \emph{haha}); (4) Replace punctuations with space; (5) Replace multiple spaces with single space. For the SMS (small) and SMS (large) datasets, we carry out an additional step of removing group messaging commands. 

We evaluate the performance of our framework against a manually annotated database of Roman Urdu variations developed by 
\citet{DBLP:conf/ictai/KhanK12}. This database, which we refer to as the `gold standard', is developed from a sample of the SMS (small) dataset. It maps each word to a unique ID representing its standard or normal form. 
There are 61,000 distinct variations in the database, which map onto 22,700 unique IDs. The number of variations differs widely for different unique IDs. For example, \textit{mahabbat} [love] has over 70 variations such as \textit{muhabaat}, \textit{muhabbat}, and \textit{mhbt}. 
The gold standard database also includes variations of English language words that are present in the dataset. 

Table \ref{tab:data} shows statistics of the datasets in comparison with the evaluation gold standard database. The ``Overlap with Gold Standard'' means the number of words in the vocabulary of a dataset that also appear in the gold standard lexicon~\cite{DBLP:conf/ictai/KhanK12}. The table also gives the number of words that appear in the gold standard \emph{and} have at least (1) one preceding and at least one following word (context size $\geq$ 1), and (2) five distinct preceding and following words in the dataset (context size $\geq$ 5). We evaluate these numbers of words for the respective datasets. The UrduPhone IDs of a dataset gives the number of distinct encodings of the evaluation words in the dataset (corresponding to the number of initial clusters). 

\begin{table}[!t]
    \caption{Datasets and gold standard database statistics \label{tab:data}}
    \begin{minipage}{\textwidth}
    {\begin{tabular}{l | r r r r}
        \toprule
        Dataset & SMS (small) & SMS (large) & CFMP & Web \\
        \midrule
        Message Count & 159,158 & 1,994,136 & 183,083 & 5,423 \\[1mm]
        Unique words & 89,692 & 366,583 & 101,395 & 21,800 \\[1mm]
        \begin{minipage}{40mm}Overlap with \\Gold Standard (OGS)\end{minipage} & 57,699 & 51,477 & 23,112 & 12,634 \\[4mm]
        \begin{minipage}{40mm}OGS and context\\ information $\geq 1$\end{minipage} & 51,133 & 49,272 & 18,516 & 9,773 \\[4mm]
        \begin{minipage}{40mm}UrduPhone IDs \\ Previous Case\end{minipage}& 11,146 & 9,738 & 4,683 & 6,171 \\[4mm]
        \begin{minipage}{40mm}OGS and \\context information $\geq 5$\end{minipage} & 12,852 & 30,856 & 1,414 & 2,479 \\[4mm]
        \begin{minipage}{40mm}UrduPhone IDs for\\ Previous Case\end{minipage} & 4,218 & 6,681 & 1,305 & 2,175 \\
        \bottomrule
    \end{tabular}}
    \end{minipage}
\end{table}

\subsection{Experimental Results and Analysis}
\label{subsec:exp}
We conduct different experiments to evaluate the performance of our clustering framework for lexical normalization of Roman Urdu. We test different combinations of features (UrduPhone, string, and/or, context) and different representations of contextual information (UrduPhone IDs or word IDs). We also establish two baseline methods for comparisons.

Table \ref{tab:exp} gives the details of each experiment's setting. Exp. 1 and 2 are baselines corresponding to segmentation using UrduPhone encoding and string similarity-based clustering (with initial random clusters equal to the number of UrduPhone segments), respectively. The remaining experiments utilize different combinations of features (string, phonetic, and context) in our clustering framework. Here, for string-based features, we used manually defined edit distance rules.\footnote{Section \ref{sec:string} presents a comparison of using automatically learned edit distance rules with manually defined rules.} The initial clustering in these experiments is given by segmentation via UrduPhone encoding. In Exp. 3 no contextual information is utilized, while in Exp. 4 and Exp. 5 the context is defined by the top-5 most frequently occurring previous and next words (context size $\geq$ 5) represented by their UrduPhone IDs and word IDs, respectively. In Exp. 2 to 5, we select the similarity threshold $t$ such that the number of discovered clusters is as close as possible to the number of actual clusters in the gold standard for each dataset. This is done to make the results comparable across different settings. During our experiments, we observed that a threshold within a range of $0.25-0.3$ was optimal for smaller datasets, including Web \& CFMP, and $0.4-0.45$ gave the best performance for larger datasets, including SMS (small) \& SMS (large). However, we also tried to find the optimum threshold value using the Nelder-Mead method (see Table \ref{tab:weights_change_exp}), which maximizes the F-Measure.

\begin{figure}[!h]
\caption{Performance results for experiments in Table \ref{tab:exp}}\label{fig:results}
\subfloat[SMS (small) dataset]{
\includegraphics[width=0.5\linewidth]{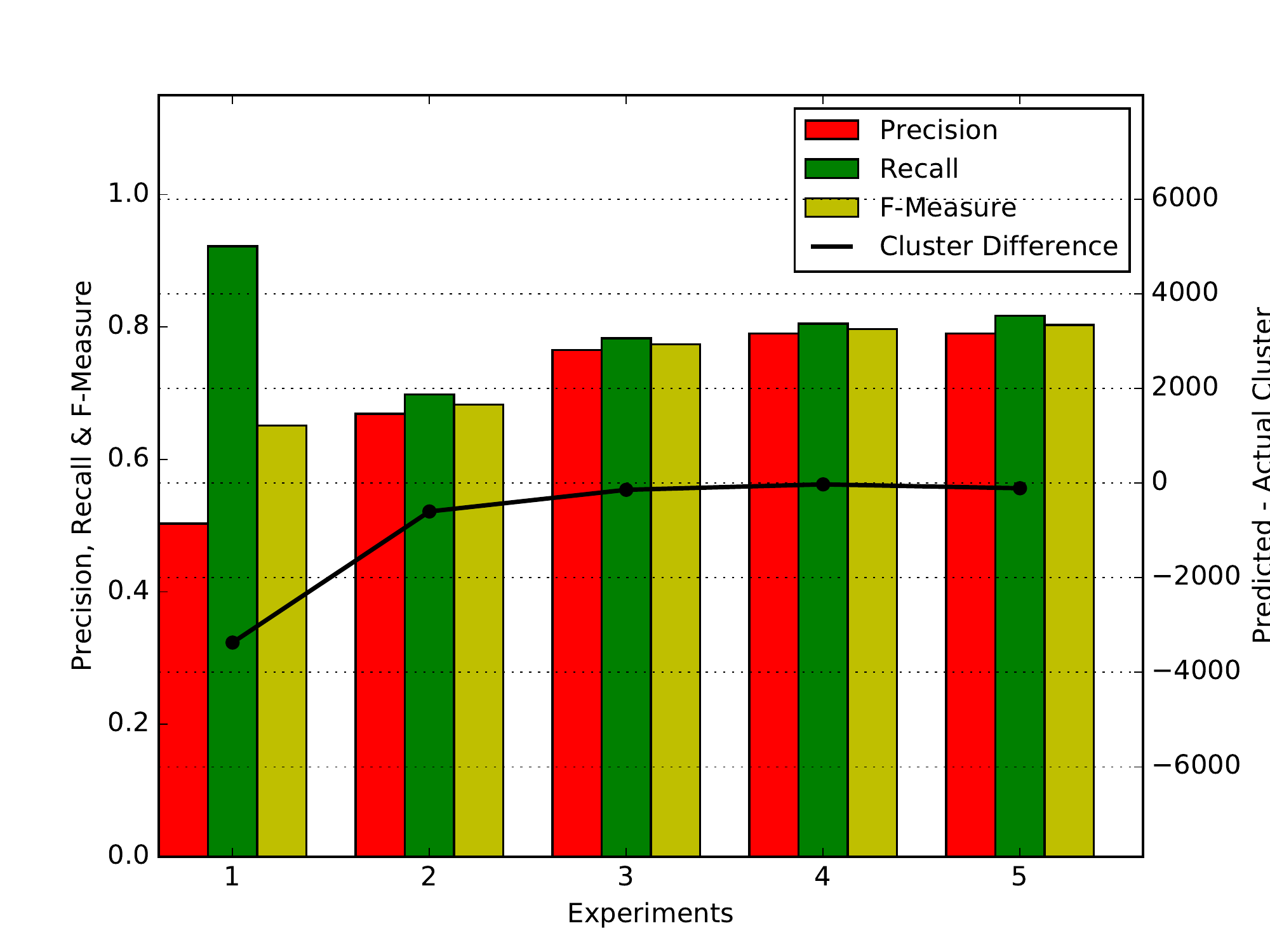}
\label{fig:sms_small}
} 
\subfloat[SMS (large) dataset]{
\includegraphics[width=0.5\linewidth]{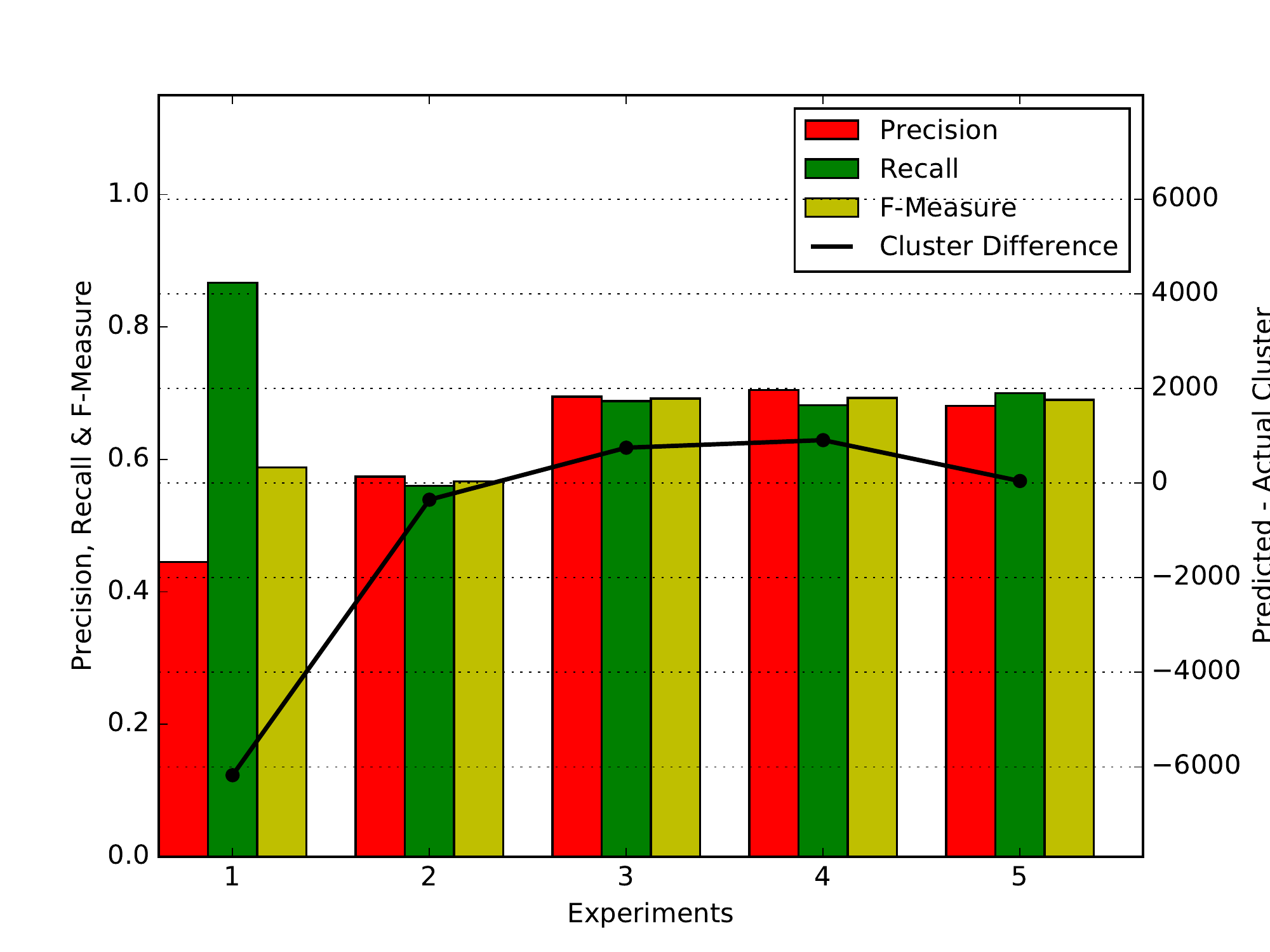}
\label{fig:sms_large}
}\\
\subfloat[CFMP dataset]{
\includegraphics[width=0.5\linewidth]{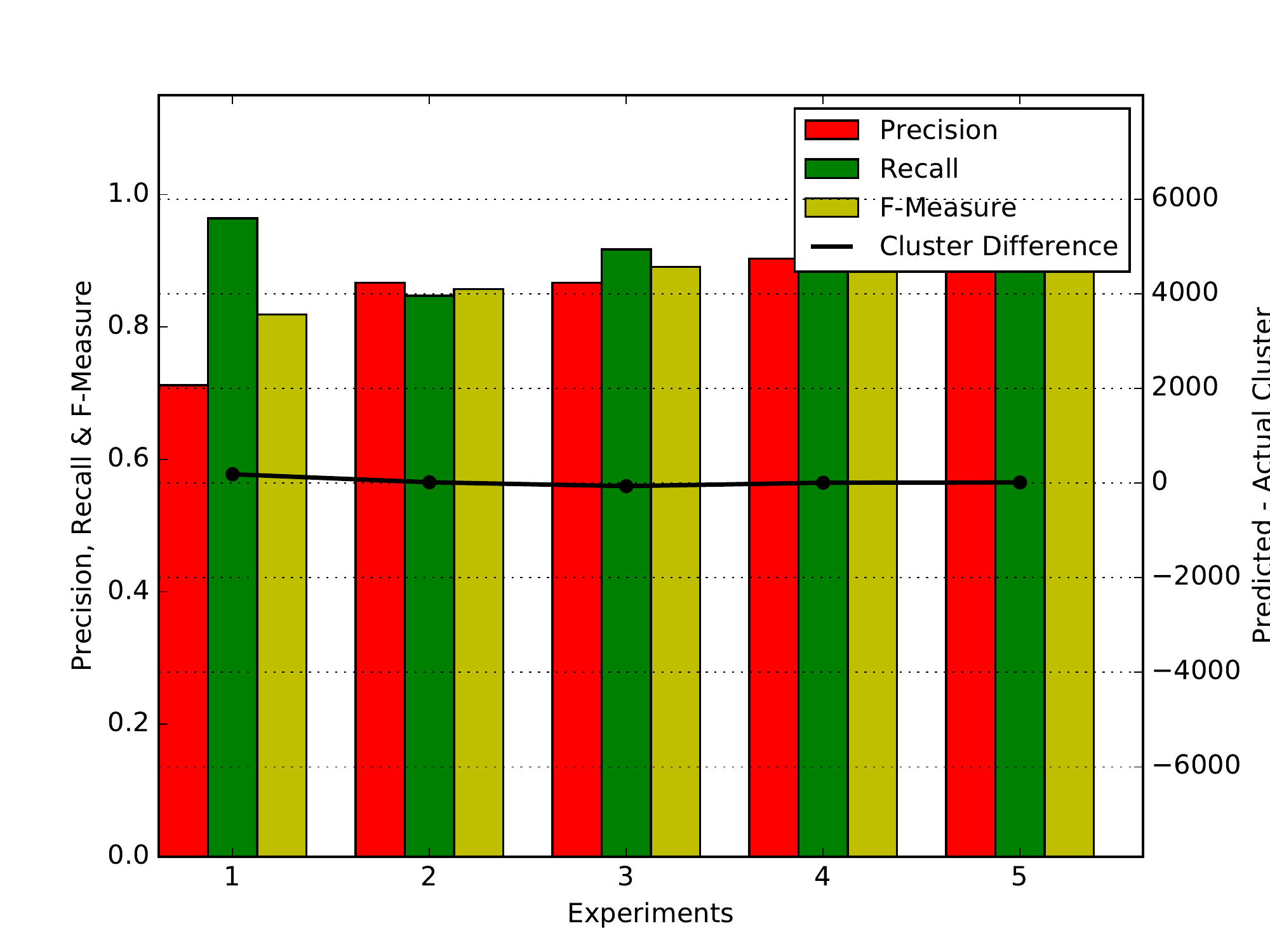}
\label{fig:cfmp}
} 
\subfloat[Web dataset]{
\includegraphics[width=0.5\linewidth]{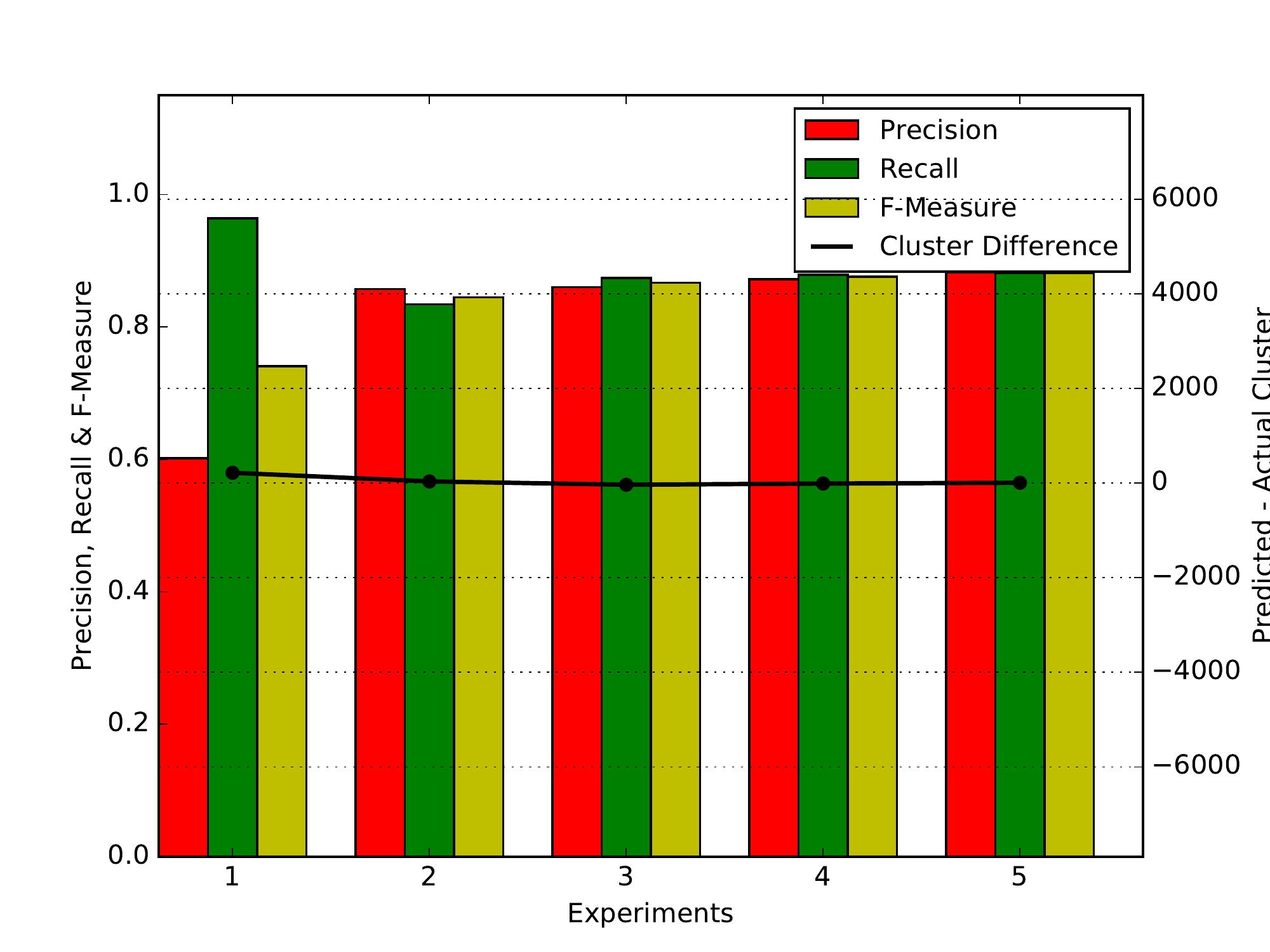}
\label{fig:web}
}
\end{figure}

Figures \ref{fig:sms_small}, \ref{fig:sms_large}, \ref{fig:cfmp}, and \ref{fig:web} show performance results on SMS (small), SMS (large), CFMP, and Web datasets, respectively. The x-axes in these figures show the experiment IDs from Table \ref{tab:exp}, while the left y-axes give the BCubed precision, recall, and F-measure, and the right y-axes describe the difference between the number of predicted and actual clusters. 

The baseline experiment of segmentation via UrduPhone encoding (Exp. 1) produces a high recall and a low precision value. This is because UrduPhone tends to group more words in a single cluster, which decreases the total number of clusters and results in an overall low F-measure. 
The second baseline of string-based clustering (Exp. 2) gives similar values for precision and recall since the average number of clusters is closer to that of the gold standard. Although the F-measure increases over Exp. 1, string-based similarity alone does not result in sound clustering.

Combining the string and phonetic features in our clustering framework (Exp. 3) results in an increase in precision and recall values as well as a marked increase in F-measure from the baselines (e.g., there is an increase of 9\% for the SMS (small) dataset,  see Fig. \ref{fig:sms_small}). When contextual information is added (via UrduPhone IDs in Exp. 4 and word IDs in Exp. 5), precision, recall, and F-measure values increase further. For example, for the SMS (small) dataset, the F-measure increases from 77.4\% to 79.7\%  (2\% gain) and from 77.4\% to 80.3\% (3\% gain) from Exp. 3 to Exp. 4 and Exp. 5, respectively. 

The higher performance values obtained for the CFMP and Web datasets (Fig. \ref{fig:cfmp} and Fig. \ref{fig:web}) are due to fewer variations in these datasets, as evidenced by their fewer numbers of unique words in comparison to the SMS datasets. 

Overall, our clustering framework using string, phonetic, and contextual features shows a significant  F-measure gain when compared to baselines Exp. 1 and Exp.  2. We obtain the best performances when we use UrduPhone and string similarity, and when the context is defined using Word IDs (Exp. 5).

\begin{table}[!t]
    \caption{Details of experiments' settings \label{tab:exp}}
    \begin{minipage}{\textwidth}
    {\begin{tabular}{c c c c c}
        \toprule
        Exp. & Initial clusters & String & Phonetic & Context\\
        \midrule
        1 & UrduPhone & \raisebox{.15ex}{\hspace{0.1em}\xmark} & \raisebox{.15ex}{\hspace{0.1em}\xmark} & -- \\
        2 & Random & \raisebox{.15ex}{\hspace{0.1em}\cmark} & \raisebox{.15ex}{\hspace{0.1em}\xmark} & -- \\
        \midrule
        3 & UrduPhone & \raisebox{.15ex}{\hspace{0.1em}\cmark} & \raisebox{.15ex}{\hspace{0.1em}\cmark} & --\\
        4 & UrduPhone & \raisebox{.15ex}{\hspace{0.1em}\cmark} & \raisebox{.15ex}{\hspace{0.1em}\cmark} & UrduPhone ID\\
        5 & UrduPhone & \raisebox{.15ex}{\hspace{0.1em}\cmark} & \raisebox{.15ex}{\hspace{0.1em}\cmark} & Word ID\\
        \bottomrule
    \end{tabular}}
    \end{minipage}
\end{table}

\subsubsection{UrduPhone}
\label{sec:urduphone_eval}
We compare UrduPhone with Soundex and its variants\footnote{We use Apache Commons Codec for DoubleMetaphone (\url{https://commons.apache.org/proper/commons-codec/apidocs/org/apache/commons/codec/language/DoubleMetaphone.html})  \& NLTK-Trainer's phonetic library (\url{https://github.com/japerk/nltk-trainer/blob/master/nltk_trainer/featx/phonetics.py}) for the remaining} for lexical normalization of Roman Urdu. 
All the phonetic encoding algorithms are used to group/segment words based on their encoding and then evaluated against the gold standard.  Table \ref{tab:sx} shows the results of this experiment on the SMS (small) dataset.

We observe that UrduPhone outperforms Soundex, Caverphone, and Metaphone while NYSIIS's F-measure is comparable to that of UrduPhone. NYSIIS produces a large number of single-word clusters (4,376 have only one word out of 6,550 groups), which negatively impacts its recall. 
UrduPhone produces fewer clusters (and fewer one-word clusters), giving high recall. This property of UrduPhone is desirable for initial clustering in our clustering framework, as Lex-Var can split them but cannot collapse them. 

We also test our clustering framework by replacing UrduPhone with NYSIIS as the phonetic algorithm. In Exp. 5 on the SMS (small) dataset, we find that the F-measure increases by only 5\% over the NYSIIS baseline (Table \ref{tab:sx}), which is lower than the F-measure achieved with UrduPhone (Fig. \ref{fig:sms_small}). 

\begin{table}[!t]
    \caption{Comparison of UrduPhone with other algorithms on the SMS (small) dataset. Single clusters are clusters with one word only. Actual clusters = 7,589\label{tab:sx}}
    \begin{minipage}{\textwidth}
    {\begin{tabular}{l | r r r r r}
        \toprule
        Algorithm & Precision & Recall & F-measure & Clusters & \begin{minipage}{5mm}Single\\ Clusters\end{minipage}\\[2mm]
        \midrule
        Soundex & 0.216 & 0.960 & 0.353 & 1,647 & 525\\[2mm]
        Metaphone & 0.468 & 0.871 & 0.601 & 3,906 & 2,061\\[2mm]
        \begin{minipage}[2cm]{35mm}Double Metaphone\\Primary Encoding\end{minipage} & 0.295 & 0.931 & 0.448 & 2388 & 1008\\[4mm]
        \begin{minipage}{35mm}Double Metaphone \\ Alternative Encoding\end{minipage} & 0.280 & 0.927 & 0.430 & 2291 & 964\\[4mm]
        Caverphone & 0.286 & 0.885 & 0.433 & 2,498 & 1,315\\
        NYSIIS & 0.584 & 0.668 & 0.623 & 6,550 & 4,376\\
        UrduPhone & 0.508 & 0.923 & \textbf{0.655} &  4,272 & 2,399\\
        \bottomrule
    \end{tabular}}
    \end{minipage}
\end{table}

\begin{figure}[h]
\centering
\caption{Effect of varying UrduPhone encoding length on SMS (small) dataset (Exp 5)}
    \label{fig:urduphone_len}
  \includegraphics[scale=0.42]{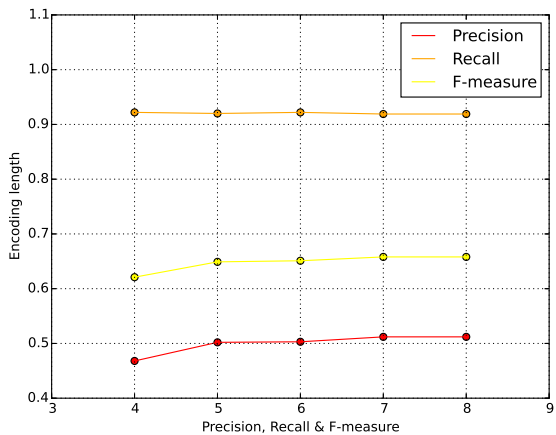}
\end{figure}

In another experiment, we 
analyze the effect of encoding length on the performance of the algorithm. We use the SMS (small) dataset to generate UrduPhone encodings of different sizes and cluster the words accordingly. Fig. \ref{fig:urduphone_len} summarizes the results.  
We see an increase in F-measure with an increase in encoding length until length seven and eight, where we achieve similar performance.

\seturdu
Table \ref{tab:soundex} defines the UrduPhone rules based on well-known techniques used for phonetic encoding schemes (dropping vowels) and on common knowledge of how people write Roman Urdu. As an additional experiment, we try to learn these rules from some datasets and use them to define our encoding scheme. We call this approach UrduPhone$_{prob}$. \citet{jiampojamarn2007} propose an alignment tool\footnote{https://github.com/letter-to-phoneme/m2m-aligner} based on the initial work of \citet{RYsed98}. Instead of mapping each grapheme to a single phoneme, their method creates a many-to-many mapping. We use an Urdu script, and Roman Urdu transliteration parallel corpus scraped from the internet.\footnote{http://www.ijunoon.com/transliteration/} Unlike the Roman Urdu words in our experiment dataset, these have more standardized spellings. We use a maximum length of two as a parameter for training the model. Our output is probabilities of Roman Urdu characters mapping to Urdu script characters or to null.
\setnone

We use the maximum probability mapping rules to define our UrduPhone$_{prob}$ encodings. We experimented with using UrduPhone$_{prob}$ as the feature in our system and also in combination with other string and context features. Table \ref{tab:exp_urduphone2} shows the results.

\begin{table}[!t]
    \caption{Experiments using UrduPhone, learning rules from Urdu-Roman Urdu transliteration corpus\label{tab:exp_urduphone2}}
    \begin{minipage}{\textwidth}
    {\begin{tabular}{l | r r r}
        \toprule
        Features & Precision & Recall & F-measure\\
        \midrule
        UrduPhone (Exp. 1) & 0.508 & 0.923 & 0.655 \\
        UrduPhone  + String + Context (Exp. 5) & 0.790 & 0.817 & 0.803 \\
        \midrule
        UrduPhone$_{prob}$ & 0.503 & 0.922 & 0.651 \\
        UrduPhone$_{prob}$ + String + Context & 0.512 & 0.919 & 0.658 \\
        \bottomrule
    \end{tabular}}
    \end{minipage}
\end{table}

\subsubsection{String-similarity}
\label{sec:ed_cost}

In section \ref{sec:string}, using the SMS (small) dataset, we compare the performance of three methods used to calculate edit distance cost -- manually defined (edist$_{man}$), automatically learned using GIZA++ (edist$_{giza}$), and automatically learned using unsupervised transliteration mining (edist$_{miner}$).\footnote{The experiment reported in previous sections used the manually defined edit distance cost, which associates cost of 1 for each insertion, deletion, and substitution operation.} 

For each word in our vocabulary, we found the 100 closest pairs, where closeness is defined by our similarity function as described in Eq. \ref{eq:similarity_function} using UrduPhone, edist$_{man}$ for the string similarity, and context of previous and next Word IDs as the feature set. We created a list of candidate word pairs by pairing every word with every other word in the cluster of 100 closest words. We take each Roman Urdu word as a sequence of Roman characters and its original Urdu script as a sequence of Urdu characters. We learn the alignment between the above two character-sequences in two different ways.  First, we apply GIZA++ and learn the alignment with the Expectation-Maximization (EM) algorithm. Second, we implemented an unsupervised transliteration mining tool, details see~\cite{sajjad2017statistical}.
Here, GIZA++ considers every word pair in the list of candidate pairs as a correct word pair to learn character alignments, whereas the transliteration mining tool penalizes the pairs that are less likely to be transliterations of each other during the training process. Since our list of candidate pairs is a mix of correct and incorrect pairs, the character alignments learned by the transliteration miner are likely to be better.
The edit distance cost for each pair of characters can be computed from character alignments as $cost(char_i,char_j) = |1-P(char_i,char_j)|$. Our string similarity function uses these edit distance costs instead of manually defined costs. 
Table \ref{tab:edist_cost} reports the results for both of these experiments using the SMS (small) dataset. The F-measure of the cost learned by the miner and GIZA++ is competitive with the manually defined cost. edist$_{giza}$ is affected by the noise in the data, which can be seen in its low precision compared to other methods. edist$_{miner}$ achieved the highest precision, though it has the lowest recall.

\begin{table}[!t]
    \caption{Varying edit distance cost for SMS (small) dataset. Learning character pair alignment probabilities\label{tab:edist_cost}}
    \begin{minipage}{\textwidth}
    {\begin{tabular}{l | r r r}
        \toprule
        String feature & Precision & Recall & F-measure\\
        \midrule
        edist$_{man}$ (Exp. 5) & 0.790 & 0.817 & 0.803\\
        \midrule
        edist$_{giza}$ & 0.786 & 0.817 & 0.802\\
        edist$_{miner}$ & 0.794 & 0.813 & 0.803\\
        \bottomrule
    \end{tabular}}
    \end{minipage}
\end{table}
\subsubsection{Context Size}
\label{sec:contexts}
The experiments presented in the previous section used a context of top-5 frequently occurring previous and next words. Here, we study the effect of varying context size on the performance of our clustering framework. Table \ref{tab:results} shows the F-measure for all experiments with two different context sizes on the SMS (small) dataset. Decreasing the minimum context list size to one increases the number of words to evaluate; therefore, results are reported for all experiments with context size between 1 and 5, even though Exp. 1 to 3 do not use contextual information. Decreasing the minimum context list size to one also explains the lower performance values for these experiments as compared to those with context size of at least 5. 

We see that context size of 1 to 5 (including words with contexts defined by at least 1 to 5 top previous/next words) is less effective in lexical normalization and sometimes even negatively impacts performance. For example, for the SMS (small) and CFMP datasets, Exp. 3 (no contextual information) performs better than Exp. 4 and Exp. 5 due to the noisy nature of shorter contexts.

\begin{table}[!t]
    \caption{Performance (F-measure) with two different context sizes. Details of the experiments are given in Table \ref{tab:exp}.\label{tab:results}}
    \begin{minipage}{\textwidth}
    {\begin{tabular}{c | l l l l}
        \toprule
        Exp. & SMS (small) & SMS (large) & CFMP & Web \\
        \midrule
        \multicolumn{5}{c}{Context Size = 5}\\
        \midrule
        1 & 0.651 & 0.588 & 0.852 & 0.831 \\
        2 & 0.683 & 0.567 & 0.857 & 0.845 \\        
        3 & 0.774 & 0.692 & 0.891 & 0.867 \\
        4 & 0.797 & \textbf{0.693} & 0.900 & 0.876 \\ 
        5 & \textbf{0.803} & 0.690 & \textbf{0.917} & \textbf{0.881} \\
        \midrule
        \multicolumn{5}{c}{Context Size = 1 to 5}\\
        \midrule
        1 & 0.593 & 0.576 & 0.616 & 0.641 \\
        2 & 0.542 & 0.537 & 0.598 & 0.756 \\        
        3 & 0.658 & 0.645 & 0.712 & 0.785 \\
        4 & 0.617 & 0.642 & 0.692 & 0.778 \\
        5 & 0.637 & 0.640 & 0.695 & 0.794 \\
        \bottomrule
    \end{tabular}}
    \end{minipage}
\end{table}

\begin{figure}[h]
\centering
\caption{Effect of varying context size on SMS (small) dataset (Exp 5)}
  \includegraphics[scale=0.42]{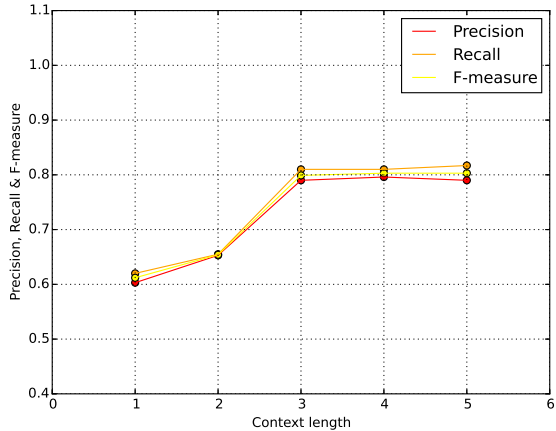}
    \label{fig:context_len}
\end{figure}

For further analysis, we carried out experiments where we changed the context length from 1 to 5; an approach that differs from the previous experiments in which we used context size $=5$ \& $\geq 1$. Fig. \ref{fig:context_len} describes the results of the tests carried out on the SMS (small) dataset. 
We see a significant increase in performance when context size changes from 2 to 3. After 3, there is a slight performance increase. The best F-measure is from a context size of 4 and 5.

\subsubsection{Parameters: Feature Weights and Clustering Threshold}
\label{sec:weights}

\paragraph{Feature Weights}

As discussed in Section~\ref{sec:featureweights}, we test the impact of changing the weights in our clustering framework (see Eq. \ref{eq:similarity_function}). We assumed that all features have equal weights in experiments presented in Section \ref{subsec:exp}. 
Then, we change the feature weights to emphasize different features. 
The increased weights caused words to break their initial UrduPhone clusters in favor of better contextual similarity, but the overall performance did not change. We tried several combinations, including using both the contexts (i.e., word IDs and UrduPhone IDs). 

Table \ref{tab:weights_change_exp} shows the performance of our clustering framework on the SMS (small) dataset with different feature weight combinations. As a comparison, we show results for Exp. 5 (context represented by word IDs only) and have the following observations with respect to F-Measure. (1) F-measure does not improve when using both word IDs and UrduPhone IDs to represent the context. (2) F-measure degrades when removing the phonetic similarity feature. (3) F-measure achieves the highest value when we set a higher weight to phonetic and contextual similarity than to string similarity.

We also use the Nelder-Mead method to maximize the F-measure by optimizing the feature weights of our similarity function in Eq. \ref{eq:similarity_function}, as well as the threshold mentioned in line 21 of Algorithm \ref{alg:lexC} on cross-validation set (see Section~\ref{sec:featureweights}). The average F-measure is slightly better than what we observed with manual selection of weights in Exp. 5 (described in Table \ref{tab:exp}).

\begin{table}[!t]
    \caption{Performance with different weights for features (Exp. 5 on SMS (small) dataset). $\alpha^P$ = Weight of phonetic feature, $\alpha^S$ = Weight of string feature, $\alpha^{C_1}$ = Weight of context using Word ID , $\alpha^{C_2}$ = Weight of context using UrduPhone ID. \label{tab:weights_change_exp}}
    \begin{minipage}{\textwidth}
    {\begin{tabular}{c l l l l}
        \toprule
        Experiment & Precision & Recall & F-measure\\
        \midrule
        Exp. 5 & 0.790 & 0.817 & 0.803\\
        \midrule
        Nelder-Mead method & 0.797 & 0.843 & 0.819\\
        \midrule
        $\alpha^P = 1.0$, $\alpha^S = 1.0$, $\alpha^{C_1} = 1.0$, $\alpha^{C_2} = 1.0$ & 0.777 & 0.814 & 0.795 \\
        $\alpha^P = 1.0$, $\alpha^S = 1.0$, $\alpha^{C_1} = 2.0$, $\alpha^{C_2} = 0.0$ & 0.784 & 0.810 & 0.797 \\
        $\alpha^P = 1.0$, $\alpha^S = 1.5$, $\alpha^{C_1} = 2.0$, $\alpha^{C_2} = 0.0$ & 0.801 & 0.812 & 0.807 \\
        $\alpha^P = 1.0$, $\alpha^S = 1.0$, $\alpha^{C_1} = 1.5$, $\alpha^{C_2} = 0.0$ & 0.801 & 0.811 & 0.806\\
        $\alpha^P = 1.5$, $\alpha^S = 1.0$, $\alpha^{C_1} = 2.0$, $\alpha^{C_2} = 0.0$ & 0.701 & 0.819 & 0.805\\
        $\alpha^P = 1.0$, $\alpha^S = 1.0$, $\alpha^{C_1} = 0.0$, $\alpha^{C_2} = 2.0$ & 0.768 & 0.781 & 0.774\\
        $\alpha^P = 1.0$, $\alpha^S = 1.0$, $\alpha^{C_1} = 2.0$, $\alpha^{C_2} = 1.5$ & 0.736 & 0.763 & 0.749\\
        $\alpha^P = 1.0$, $\alpha^S = 1.0$, $\alpha^{C_1} = 1.5$, $\alpha^{C_2} = 0.5$ & 0.793 & 0.809 & 0.801\\
        $\alpha^P = 0.0$, $\alpha^S = 1.0$, $\alpha^{C_1} = 1.0$, $\alpha^{C_2} = 0.0$ & 0.754 & 0.758 & 0.756\\            
        $\alpha^P = 0.0$, $\alpha^S = 1.0$, $\alpha^{C_1} = 1.5$, $\alpha^{C_2} = 0.0$ & 0.710 & 0.726 & 0.717\\
        $\alpha^P = 1.5$, $\alpha^S = 1.0$, $\alpha^{C_1} = 1.0$, $\alpha^{C_2} = 0.0$ & 0.802 & 0.811 & 0.807\\
        $\alpha^P = 1.5$, $\alpha^S = 1.0$, $\alpha^{C_1} = 1.5$, $\alpha^{C_2} = 0.0$ & 0.804 & 0.813 & 0.808\\
        $\alpha^P = 2.0$, $\alpha^S = 1.0$, $\alpha^{C_1} = 2.0$, $\alpha^{C_2} = 0.0$ & 0.813 & 0.809 & 0.811\\
        $\alpha^P = 2.0$, $\alpha^S = 1.5$, $\alpha^{C_1} = 2.0$, $\alpha^{C_2} = 0.0$ & 0.791 & 0.815 & 0.803\\
        \bottomrule
    \end{tabular}}
    \end{minipage}
\end{table}

\paragraph{Clustering Threshold}

\begin{figure}[h]
\centering
\caption{Effect of varying threshold $t$ on SMS (small) dataset (Exp 5)}
  \includegraphics[scale=0.42]{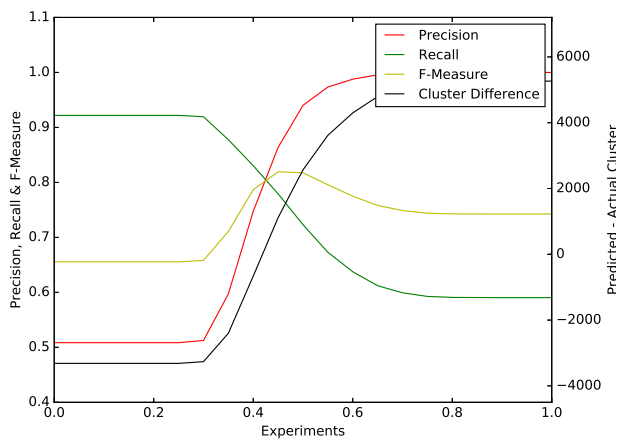}
    \label{fig:web_t}
\end{figure}

We analyze the performance of Exp. 5 (best setting) for the SMS (small) dataset with varying threshold $t$ (Fig. \ref{fig:web_t}).
The value of $t$ controls the number of clusters smoothly, and precision increases with this number while F-measure reaches a peak when the number of predicted groups is close to that of the gold standard.

\subsubsection{Comparison with Other Clustering Methods and Variations}
\label{sec:lexc_vs_others}

In addition to our k-medoids based Lex-Var clustering method, we propose using agglomerative hierarchical clustering (Hierarchical Lex-Var) as our clustering framework for lexical normalization. To reduce the search complexity at each merge decision, we form (once) and search within the ten most similar words for each word (neighborhood). At each merge decision, we merge the two most similar words and/or groups (if either word is part of a group) in their respective neighborhoods. Algorithm \ref{alg:lexHeirarchical} describes the Hierarchical Lex-Var Clustering algorithm. We tested with a neighborhood size of 10 and 100. The results are mentioned in Table \ref{tab:exp_heirarchical}.

Hierarchical Lex-Var, when used instead of Lex-Var, results in slightly better performance. However, it is significantly slower than Lex-Var. Even with our neighborhood-based optimization, hierarchical clustering takes hours to converge, while our Lex-Var algorithm converges in minutes when processing the SMS (small) dataset. 

\begin{table}[!t]
    \caption{Performance of Hierarchical Lex-Var on SMS (small) dataset.\label{tab:exp_heirarchical}}
    \begin{minipage}{\textwidth}
    {\begin{tabular}{c | l l l l}
        \toprule
        Experiment & Precision & Recall & F-measure\\
        \midrule
        Exp. 5 & 0.790 & 0.817 & 0.803\\
        Nelder-Mead method & 0.797 & 0.843 & 0.819\\
        \midrule
        Neighborhood$=10$ & 0.793 & 0.837 & 0.815\\
        Neighborhood$=100$ & 0.771 & 0.849 & 0.808\\
        \bottomrule
    \end{tabular}}
    \end{minipage}
\end{table}

\begin{algorithm2e}[!h]
  \scriptsize
  \NoCaptionOfAlgo
\KwIn{$\mathcal{W} =\{w_1,w_2,\ldots ,w_N\}$ (words), $t$ (similarity threshold), $K$ (neighborhood size)}
\KwOut{$\mathcal{L} = \{\ell_1,\ell_2,\ldots ,\ell_{K}\}$ (predicted clusters) }
$\mathcal{L} = \mathcal{W}$;\\
\tcc{Create a similarity matrix}
\For{$\forall\: w_i \: \in \: \mathcal{W}$}{
    \For{$\forall \: w_j \: \in \: \mathcal{W}\setminus w_i$}{  
        $Sim\_matrix_{i,j} = S(w_i,w_j)$;
    }
}
\tcc{Create $K$ sized neighborhoods for each word}
$\mathcal{N} = \{\}$;\\
\For{$\forall \: w_i \: \in \: \mathcal{W}$}{
        $n_i = get\_max\_k(Sim\_matrix_{i,},K)$;\tcp*[f]{Get $K$ most similar words to $w_i$}\\
        $\mathcal{N} = \mathcal{N}\cup \{n_i\}$;
}
\Repeat{stop condition Satisfied}{
    \tcc{Assign word to clusters }
    \For{$\forall ~ w_i ~ \in ~ \mathcal{W}$}{ 
        $closest =$ null;\\
        $maxSim = 0$;\\
        \For{$\forall ~ w_j ~ \in ~ \mathcal{N}_i$}{
            \If{$S(w_i,n_j)> t$ \normalfont{and} $S(w_i,n_j)>maxSim$}{
                $maxSim = S(w_i,n_j)$;\\
                $closest = n_j$;\\
            }
        }
        \If(\tcp*[f]{Move word $w_i$ to cluster $\ell_{j}$}){$closest ~ != ~ $\normalfont{null}}{
             $\ell_j = \ell_j \cup \{w_i\} \mid closest \in \ell_j$\;}
    }
}
\caption{Algorithm \ref{alg:lexHeirarchical}: Hierarchical Lex-Var}\label{alg:lexHeirarchical}
\end{algorithm2e}

Additionally, we compare our clustering framework with other clustering methods as independent approaches. We also test with variations in similarity features of our clustering framework. 
We report the following experiments: 
\begin{enumerate}
\item Rule-based transliteration: Each word in the vocabulary was transliterated based on the method by \citet{ahmed2009roman}. The final words were mapped to an Urdu word dictionary of around 150,000 words.\footnote{\url{https://raw.githubusercontent.com/urduhack/urdu-words/master/words.txt}} Each Urdu word acted as a cluster label.
\item Brown clustering: Brown clustering is a hierarchical clustering method for grouping words based on their contextual usage in a corpus  \citep{Brown:1992:CNG:176313.176316}. We use this as an independent approach for the lexical normalization of Roman Urdu. 
\item Word2Vec clustering: Word2Vec represents words appearing in a corpus by fixed-length vectors that capture their contextual usage in the corpus \citep{DBLP:journals/corr/MikolovSCCD13}. The Word2Vec model is generated using the gensim\footnote{\url{https://github.com/RaRe-Technologies/gensim}} python package to learn vectors for each Roman word. For learning the word vectors, we used the minimum count of 5, dimension size of 100, and 10 iterations.  Words are clustered using K-Means clustering on word vectors, and we report the performance for lexical normalization of Roman Urdu.
\item 2-skip-1-grams: In our clustering framework for lexical normalization, we use the 2-skip-1-gram approach with Jaccard coefficient \citep{jin:2015:WNUT} to compute string similarity (rather than our string similarity function (Eq. \ref{eq:string_similarity})). Algorithm \ref{alg:skipgram} shows the 2-skip-1-gram algorithm.
\item 2-skip-1-gram + string feature: We use both 2-skip-1-gram and our string similarity functions for computing string similarity in our clustering framework for lexical normalization. 
\item `h' omitted UrduPhone: We use a modified version of UrduPhone in our clustering framework for lexical normalization. The modified version discards aspirated characters in the encoding. For example, encoding for \emph{mujhay} [me] becomes identical to that for \emph{mujay} [me] to handle 'h' omission. 
\item Word2Vec Vectors (50): We generate Word2Vec vecctors of size 50. We use the cosine similarity of these vectors instead of the contextual similarity described in Equation \ref{eq:rank_corr}.
\item  Word2Vec Vectors (100): We increase the size of Word2Vec vectors to 100.
\item Word2Vec Words: Word2Vec vectors are used to find the ten most similar words for each word. These neighboring words define the context of each word, and contextual similarity is computed using Eq. \ref{eq:rank_corr}. We use our clustering framework for lexical normalization. 
\item Word IDs + Word2Vec Words: We use two contextual features: top-5 frequently occurring previous/next words represented by word IDs (like in Exp. 5) and top-10 most-similar words according to Word2Vec (as above). 
\end{enumerate} 

Table \ref{tab:extra_exp} summarizes the results. Experiment 1 is a rule-based lexical normalization method. Experiments 2 and 3 are independent clustering methods for lexical normalization. We also modify string features (experiments 4 and 5),  phonetic features (experiment 6), and contextual features (experiments 7, 8, 9, and 10), respectively, in our clustering framework.

    
\begin{algorithm2e}[!h]
  \scriptsize
  \NoCaptionOfAlgo
\KwIn{$w_i,w_j$ (pair of words)}
\KwOut{$\sigma_{ij}^{G}$ (2-skip-1-gram similarity)}
$m = length(word_i)$;\\
$A = \phi$;\\
\For{$k \in 1,m-2$}{
    $X = \{word_i[k]\}, \{word_i[k+2]\}$;\\
    $A = A\cup X$;
}
$n = length(word_j)$;\\
$B = \phi$;\\
\For{$l \in \{1\cdots n-2\}$}{
    $Y = \{word_j[l]\}, \{word_j[l+2]\}$;\\
    $B = B\cup Y$;
}
$\sigma_{ij}^{G} = \frac{|A\cap B|}{|A\cup B|}$;
\caption{Algorithm \ref{alg:skipgram}: 2-skip-1-gram}
\label{alg:skipgram}
\end{algorithm2e}

We can make the following observations from these experiments. (1) Rule-based transliteration performs slightly lower than our clustering method (2) Brown clustering and Word2Vec clustering are unsuitable for lexical normalization as evidenced by their poor performance. (3) Word2Vec-based context (either Word2Vec vectors or similar words) and 2-skip-1-gram-based string features do not outperform our context and string features. One possible reason for the low performance of Brown clustering and Word2Vec could be the small size of the training data. These algorithms require a huge amount of data to learn.

\begin{table}[!t]
    \caption{Performance of other  clustering methods and variations in our framework on SMS (small) dataset.\label{tab:extra_exp}}
    \begin{minipage}{\textwidth}
    {\begin{tabular}{c | l | l l l l}
        \toprule
        & Experiment & Precision & Recall & F-measure\\
        \midrule
        & Rule-based (\cite{ahmed2009roman}) & 0.833 & 0.765 & 0.797 \\
        \multirow{2}{*}{Other methods}
         & Brown clustering & 0.024 & 0.447 & 0.046\\
         & Word2Vec clustering & 0.350 & 0.221 & 0.271\\
        \midrule
        \multirow{7}{*}{Additional features}
        & 2-skip-1-gram & 0.782 & 0.810 & 0.796\\
        & 2-skip-1-gram + String feature & 0.791 & 0.799 & 0.795\\
        & 'h' omitted UrduPhone & 0.796 & 0.808 & 0.802\\ 
        & Word2Vec Vectors (50) & 0.782 & 0.802 & 0.792\\
        & Word2Vec Vectors (100) & 0.795 & 0.803 & 0.799\\    
        & Word2Vec Words & 0.777 & 0.779 & 0.778\\
        & Word IDs + Word2Vec Words & 0.780 & 0.808 & 0.793\\
        \bottomrule
    \end{tabular}}
    \end{minipage}
\end{table}

\subsubsection{Lexical Normalization of English Text}
To test the robustness of our dataset for other languages, we experimented with an English dataset provided by \citet{leon13twitter} and used in the W-NUT 2015 task.\footnote{\url{https://noisy-text.github.io/norm-shared-task.html}} The gold standard we used is the lexical normalization dictionary provided by the University of Melbourne.\footnote{Available on the W-NUT 2015 website} The dataset has more than 160,000 messages containing 60,000 unique words. After pre-processing (the same pre-processing steps as for the Roman Urdu datasets), we get a 2,700 word-overlap with the gold standard. For the phonetic encoding, we tested with Soundex and UrduPhone. 

Table \ref{tab:lang_exp} summarizes the results along with the best results for the Roman Urdu dataset from Table \ref{tab:exp}. We observe an F-measure of more than 90\% with both encoding schemes, with UrduPhone performing better than Soundex. This difference in performance is presumably due to the extended encoding size in UrduPhone, which makes it possible to keep more information about the original word. 

\begin{table}[!t]
\caption{Performance of Lex-Var on English dataset. We used Soundex \& UrduPhone encodings as phonetic features \label{tab:lang_exp}}
    \begin{minipage}{\textwidth}
    {\begin{tabular}{c | l l l l l}
        \toprule
        Language & Phonetic Encoding & Precision & Recall & F-measure\\
        \midrule
        Roman Urdu (Exp. 5) & UrduPhone & 0.790 & 0.817 & 0.803 \\
        \midrule
        English & Soundex & 0.950 & 0.948 & 0.949 \\
        English & UrduPhone & 0.967 & 0.961 & 0.965 \\
        \bottomrule
    \end{tabular}}
    \end{minipage}
\end{table}

\subsection{Error Analysis}
\label{sec:error_analysis}
To gain a better understanding of our clustering framework, we analyze the output of different experiments with examples of correct and incorrect lexical normalization. While lexical normalization based on UrduPhone mappings (Exp. 1) is a good starting point for finding word variations, it produces some erroneous groupings. We summarize these groupings as follows:

\begin{enumerate}
  \item \label{result:group1_1} Words that differ only in their vowels are in the same cluster:
  \begin{itemize}
    \item \emph{takiya} [pillow], \emph{tikka} [grilled meat], \emph{take}
    \item \emph{khalish} [pain], \emph{khuloos} [sincerity]
    \item \emph{baatain} [conversations], \emph{button}
    \item \emph{doosra} [another], \emph{desire} 
    \item \emph{separate}, \emph{spirit}, \emph{support}
  \end{itemize}
  \item \label{result:group1_2} Same words having different consonants map to different groups:
  \begin{itemize}
    \item \emph{mujhse}, \emph{mujse} meaning [from me]
    \item \emph{kuto}, \emph{kuton} meaning [dogs]
    \item \emph{whose}, \emph{whoze}
    \item \emph{skool}, \emph{school}
  \end{itemize}
  \item \label{result:group1_3} Words whose abbreviations or short forms do not have the same UrduPhone mapping:
  \begin{itemize}
    \item \emph{government}, \emph{govt} 
    \item \emph{private}, \emph{pvt}
    \item \emph{because}, \emph{coz}
    \item \emph{forward}, \emph{fwd}
  \end{itemize}
\end{enumerate}

Exp. 4 and Exp. 5 can separate words initially clustered incorrectly (group \ref{result:group1_1}) (e.g., \emph{baatain} [conversations] and \emph{button}, \emph{spirit} and \emph{support}) due to contextual information and similarity differentiating the variations. Despite using phonetic variations in combination with contextual feature we see incorrect clusterings in the two experiments. We can divide these inaccuracies into several groups.
\begin{enumerate}
\item \label{result:group4_1} Words that have different UrduPhone mappings but are in fact the same. These are not clustered in the final outcome. 
  \begin{itemize}
  \item {[\emph{mujy}]} and {[\emph{mujhy}]} meaning [me]
  \item {[\emph{oper}]} and {[\emph{uper}]} meaning [up]
  \item {[\emph{prob}]} and {[\emph{problem}]}
  \item {[\emph{mornin}]} and {[\emph{morng}]}
  \item {[\emph{number}]} and {[\emph{numbers}]}
  \item {[\emph{please}]} and {[\emph{plx,plz}]}
  \end{itemize}
\item \label{result:group4_2} Words that have the same UrduPhone mapping and are lexical variants but are not clustered in the same group:
  \begin{itemize}
  \item {[\emph{tareeka}]} and {[\emph{tareka}]} meaning [way]
  \item {[\emph{zamaane}]} and {[\emph{zamany}]} meaning [times]
  \item {[\emph{msg}]} and {[\emph{message}]}
  \item {[\emph{morng}]} and {[\emph{morning}]}
  \item {[\emph{cmplete,complet,complete}]} and {[\emph{cmplt}]}
  \end{itemize}
  \item \label{result:group4_3} Words that are different but have the same UrduPhone mapping and are clustered together:
  \begin{itemize}
  \item \emph{maalik} [owner], \emph{malika} [queen], \emph{malaika} [angels]
  \item \emph{nishaan} [vestige], \emph{nishana} [target]
  \item \emph{tareka} [way], \emph{tariq} [a common name meaning 'a night visitor']
  \item \emph{what}, \emph{white}
  \item \emph{waiter}, \emph{water}
  \end{itemize}
\end{enumerate}

A closer look at the examples reveals that some words that have the same UrduPhone mapping and should cluster together are found in separate groups (group \ref{result:group4_2}). This result is due to low context similarity between the words, which causes them not to group (e.g., \emph{tareeka} and \emph{tareka} meaning [way] have a contextual similarity of 0.23, even though they have the same UrduPhone mapping).

Another prominent issue is that words in separate clusters in UrduPhone remain separated in the output of Exp. 4 and Exp. 5 (groups \ref{result:group4_2} and \ref{result:group4_3}). This observation highlights the point that our experiments do not perform well at handeling abbreviations (e.g., \emph{prob} and \emph{problem}), plurals (e.g., \emph{number} and \emph{numbers}), and some phonetic substitutes (e.g., \emph{please} and \emph{plx}). Our framework separates Roman Urdu words that can be written with an additional consonant (e.g., \emph{mujy} and \emph{mujhy} meaning [me]). It also maps words that start with a different vowel (e.g., \emph{oper} and \emph{uper} meaning [up]). 

To tackle the issue of low contextual similarity not overcoming the difference in UrduPhone mapping, we doubled the weight assigned to the context feature. This adjustment produces almost no change in overall performance when compared to standard (Exp. 4 and Exp. 5). However, this adjustment causes more words with different UrduPhone mappings to be clustered together, usually incorrectly:
\begin{itemize}
\setlength\itemsep{0.1em}
\item \emph{acha} [okay], \emph{nahaya} [bathe], \emph{sucha} [truthful]
\item \emph{maalom} [know], \emph{manzor} [approve]
\item \emph{chalang} [jump], \emph{thapar} [slap]
\item \emph{darzi} [tailor], \emph{pathar} [stone]
\item \emph{azmaya} [to try], \emph{sharminda} [ashamed]
\end{itemize}
Furthermore, as the same UrduPhone mappings do not restrict the clusters, this variation produces interesting combinations. The words in the groups below, although not lexical variants of each other, have strong contextual similarity and sometimes can even be replaced (for the other) in the sentence.
\begin{itemize}
\setlength\itemsep{0.1em}
\item \emph{admi} [man], \emph{larkay} [boys], \emph{larki} [girl]
\item \emph{kufr} [to unbelieve in God], \emph{shirk} [to associate partners with God]
\item \emph{shak} [suspicion], \emph{yaqeen} [certainty]
\item \emph{loves}, \emph{likes}
\item \emph{private}, \emph{pvt}
\item \emph{cud}, \emph{may}, \emph{would}
\item \emph{tue}, \emph{tuesday}, \emph{wed}    
\item \emph{blocked}, \emph{kicked}
\item \emph{gov}, \emph{government}
\end{itemize}

\section{Previous Work} 
\label{sec:previousWork}

Normalization of informal text messages and tweets has been a research topic of interest
\citep{Sproat99normalizationof,kaufmann2010syntactic,clark2011text,wei2011exploring,DBLPconftsdPintoAAGLJ12,ling2013paraphrasing,sidarenka2013rule,roylexicon,chrupala:2014:P14-2,desai2015normalization}, with the vast majority of the work limited to English and other resource-rich languages. 
Our work focuses on Roman Urdu, an under-resourced language, that does not have a gold standard corpus with standard word forms. We restrict our task to finding lexical variations in informal text, a challenging problem because every word is a possible variation of every other word in the corpus. Additionally, the spelling variation problem of Roman Urdu inherits inconsistencies that occur due to the transliteration of Urdu words from Perso-Arabic script to Roman script. In our work, we model these inconsistencies separately and in combination with other features.

Researchers have used phonetic, string, and contextual knowledge to find lexical variations in informal text.\footnote{Spelling correction is also considered as a variant of text normalization  \citep{Damerau:1964:TCD:363958.363994,Tahira2004,fossati2007mixed}. Here, we limit ourselves to the previous work on short text normalization.}
\citet{DBLPconftsdPintoAAGLJ12,han2012automatically,zhang2015normalization}  used phonetic-based methods to find lexical variations. 

\citet{contractor2010unsupervised} used string edit distance based on the longest common subsequence ratio and edit distance of Consonant Skeletons \citep{prochasson07language} of the IV-OOV words. \citet{gouws2011unsupervised} used a sizable English corpus to extract candidate lexical variations and re-score them based on lexical similarity. We also use lexical similarity as a feature in our clustering framework but do not have a reference to a Roman Urdu corpus with standard word forms. \citet{jin:2015:WNUT} also generated an OOV-IV list by using the Jaccard Index \citep{levandowsky1971} between \textit{k}-skip-\textit{n}-grams of string \textit{s} and standard word forms. As we do not have these in Roman Urdu, we consider every word as a possible lexical variation of every other word in the corpus. 
Similar to \citet{jin:2015:WNUT}, we use k-skip-n-grams in our additional experiments and find that they perform slightly worse than our algorithm.
\citet{chrupala:2014:P14-2} used Conditional Random Field \citep{Lafferty01conditionalrandom} to learn the sequence of edits from labeled data.

\citet{han2012automatically}  used word similarity and word context to enhance performance by initially extracting OOV (out-of-vocabulary) -- IV (in-vocabulary) pairs using contextual similarity and then re-ranking them based on string and phonetic distances. In contrast, we define a similarity function that considers all three features together to find lexical variations of a word.
Unlike previous approaches, we have a small corpus from which to extract contextually similar word pairs. Also, there is no standard Roman Urdu dictionary that can be used to annotate words as either IV or OOV.
\citet{li-liu:2014:P14-3} defined similarity measure as a combination of the longest common subsequence, term frequency, and inner product of word embeddings. We use the longest common subsequence as part of the string similarity feature. In our additional experiments, we test with a cosine similarity of word embeddings (Table \ref{tab:extra_exp}).
\citet{li-liu:2014:P14-3} used a combination of string similarity and vector-based similarity to generate a candidate list, which was re-ranked using a character-level machine translation model \citep{pennell2011character} and Jazzy Spell Checker,\footnote{\url{http://jazzy.sourceforge.net/}} etc.
\citet{Yang-alog-linear} used an unsupervised approach that learns string edit distance, lexical, and contextual features using a log-linear model and sequential Monte Carlo approximation.

\citet{Singh18,Bertaglia17} used word embeddings to find similar standard and non-standard words for text normalization.
\citet{chrupala:2014:P14-2} used character-level neural text embeddings \citep{DBLP:journals/corr/Chrupala13} as added information from unlabeled data for better performance. \citet{Sridhar2014} used deep neural networks to learn distributed word representations.  We experimented with word embeddings as a feature in our similarity measure in the supplementary experiments Table \ref{tab:extra_exp}.


%
\citet{Hassan-socialtext} used a 5-gram language model to create a contextual similarity lattice and applied Markov random walk for lexicon generation. Their approach uses a linear combination of contextual feature and string similarity (longest common subsequence ratio and edit distance), which is very similar to our approach. However, unlike \citet{Hassan-socialtext}, we assume that every Roman Urdu word is a noisy word and thus can not separate nodes of the graph into standard and non-standard forms. \citet{Sproat17} used a recurrent neural network to normalize text. \citet{pennell2011character,li-liu:2014:P14-3} used a character-level machine translation system for the normalization task. \citet{Massimo18} used an encoder-decoder architecture where different levels of granularity were used for the target-side language model, e.g. characters and words. \citet{wang-ng:2013:NAACL-HLT} used a beam-search decoder with integrated normalization operations such as missing word recovery and punctuation correction to normalize non-standard words. Our work, however, is limited to grouping the lexical variations of Roman Urdu words.  However, we do not have any labeled data or parallel data available to build such a translation system. Our proposed method is robust since it learns from user data, and it groups abbreviations and their complete forms together in one cluster. 

%
\citet{Almeida:2016:TNS:2989323.2989716} used a standard English dictionary and an informal English dictionary to normalize words to their root forms. In our case, we do not use a standard dictionary as one does not exist for Roman Urdu words. 
\citet{ling2013paraphrasing} automatically learned normalization rules using a parallel corpus of informal text.
\citet{Irvine2012} used manually prepared training data to build an automatic normalization system for the Roman Urdu script. Unlike \citet{Irvine2012}, we propose an unsupervised approach, which does not require labeled data. Additionally, our approach to the Roman Urdu normalization problem does not require us to have a corresponding Urdu script form for each Roman word.

\paragraph{Phonetic encoding schemes}
There have been several sound-based encoding schemes used in the literature to group similar sounding words together. Here, we summarized a few of the schemes in the context of lexical normalization.

The Soundex algorithm \citep{soundexbook,DBLP:journals/csur/HallD80} encodes the first letter and the following three consonants of a word with consonants having a similar place of articulation sharing the same code. The NYSIIS method \citep{NYSIIS}, designed by the New York Police for American names, employ more sophisticated encoding rules based on multi-character n-grams and relative vowel positioning.
The Metaphone algorithm \citep{clm/philips90}, developed in 1990 as a Soundex variant, incorporates English pronunciation rules for phonetic encoding of words. Other, more-recent variations include Caverphone \citep{DBLP:reference/dataware/2009} and Double Metaphone;\footnote{\url{http://en.wikipedia.org/wiki/Metaphone}} they include complex grammatical rules for phonetic encoding of words. The Double Metaphone algorithm also differs from others in that it generates up to two encodings for each word -- one reflects the basic version of the word's pronunciation, and the other reflects an alternative pronunciation based on other languages. This is particularly useful when comparing foreign names with their anglicized versions.  For example, the names \emph{Catherine} and \emph{Katrina} have a common code \emph{KTRN}. Previous algorithms like Metaphone and Soundex do not provide such a capability. 

Most of these schemes are designed for English and European languages and are not sufficiently expressive, especially for lexical normalization or when applied to another family of languages.

We propose a method to find lexical variations in Roman Urdu that uses string edit distance like \citet{contractor2010unsupervised}, sound-based encoding like \citet{DBLPconftsdPintoAAGLJ12}, and contextual information like \citet{han2012automatically} combined in a discriminative framework. In contrast to previous work, our method does not use a resource of standard word forms to find lexical variations.

\section{Conclusion and future work}
\label{sec:conclusion}
Roman Urdu is a transliterated form of the Urdu language written in Roman script, used in informal communication in social media and SMS texts. It does not have a standard lexicon, which results in an extensive use of lexical variations that hamper automatic processing. Our framework for lexical normalization of Roman Urdu is an unsupervised model meant to address this important issue. Our clustering framework incorporates customized phonetic encoding, string-based matching, and contextual similarity. 
We conducted an extensive evaluation of our framework on four real-world datasets. We used manually generated gold standard 
containing Roman Urdu lexical variations with their standard forms \citep{DBLP:conf/ictai/KhanK12}.
We show that our framework effectively discovers lexical variations in Roman Urdu corpora with significant improvement over the baseline methods.

Our work brings us one step closer to automatically generating a normalized Roman Urdu corpus. We can cluster spelling variations of a word and then map them to the most frequent form, and can use this corpus to develop NLP applications. In the future, we would like to extrinsically evaluate our normalization procedure on several NLP tasks, such as POS tagging and machine translation.

\section{Acknowledgement}
This research was partially funded by the National Science Foundation (NSF) Award
No. 1747728 and the National Science Foundation of China (NSFC) Award No.
61672524.

\newcommand{\newblock}{}
\bibliographystyle{apalike}
\bibliography{nle}

\begin{thebibliography}{}

\bibitem[Ahmed, 2009]{ahmed2009roman}
Ahmed, T. (2009).
\newblock Roman to {U}rdu transliteration using wordlist.
\newblock In {\em Proceedings of the Conference on Language and Technology},
  Poznan, Poland.

\bibitem[Almeida et~al., 2016]{Almeida:2016:TNS:2989323.2989716}
Almeida, T.~A., Silva, T.~P., Santos, I., and G\'{o}mez~Hidalgo, J.~M. (2016).
\newblock Text normalization and semantic indexing to enhance instant messaging
  and {SMS} spam filtering.
\newblock {\em Knowledge-Based Systems}, 108(C):25--32.

\bibitem[Bagga and Baldwin, 1998]{bCubed98}
Bagga, A. and Baldwin, B. (1998).
\newblock Algorithms for scoring coreference chains.
\newblock In {\em Proceedings of the 1st International Conference on Language
  Resources and Evaluation Workshop on Linguistics Coreference}, pages
  563--566, Granada, Spain.

\bibitem[Brown et~al., 1992]{Brown:1992:CNG:176313.176316}
Brown, P.~F., deSouza, P.~V., Mercer, R.~L., Pietra, V. J.~D., and Lai, J.~C.
  (1992).
\newblock Class-based n-gram models of natural language.
\newblock {\em Computational Linguistics}, 18(4):467--479.

\bibitem[Chrupa{\l}a, 2013]{DBLP:journals/corr/Chrupala13}
Chrupa{\l}a, G. (2013).
\newblock Text segmentation with character-level text embeddings.
\newblock In {\em Proceedings of the International Conference on Machine
  Learning: Workshop on Deep Learning for Audio, Speech and Language
  Processing}, Atlanta, Georgia, USA.

\bibitem[Chrupa{\l}a, 2014]{chrupala:2014:P14-2}
Chrupa{\l}a, G. (2014).
\newblock Normalizing tweets with edit scripts and recurrent neural embeddings.
\newblock In {\em Proceedings of the 52nd Annual Meeting of the Association for
  Computational Linguistics: Short Papers}, pages 680--686, Baltimore,
  Maryland, USA. Association for Computational Linguistics.

\bibitem[Clark and Araki, 2011]{clark2011text}
Clark, E. and Araki, K. (2011).
\newblock Text normalization in social media: Progress, problems and
  applications for a pre-processing system of casual english.
\newblock {\em Procedia-Social and Behavioral Sciences}, 27:2--11.

\bibitem[Contractor et~al., 2010]{contractor2010unsupervised}
Contractor, D., Faruquie, T.~A., and Subramaniam, L.~V. (2010).
\newblock Unsupervised cleansing of noisy text.
\newblock In {\em Proceedings of the 23rd International Conference on
  Computational Linguistics: Poster}, pages 189--196, Beijing, China.
  Association for Computational Linguistics.

\bibitem[Costa~Bertaglia and Volpe~Nunes, 2016]{Bertaglia17}
Costa~Bertaglia, T.~F. and Volpe~Nunes, M. d.~G. (2016).
\newblock Exploring word embeddings for unsupervised textual user-generated
  content normalization.
\newblock In {\em Proceedings of the 2nd Workshop on Noisy User-generated
  Text}, pages 112--120, Osaka, Japan.

\bibitem[Damerau, 1964]{Damerau:1964:TCD:363958.363994}
Damerau, F.~J. (1964).
\newblock A technique for computer detection and correction of spelling errors.
\newblock {\em Communications of the Association for Computing Machinery},
  7(3).

\bibitem[Derczynski et~al., 2013]{leon13twitter}
Derczynski, L., Ritter, A., Clark, S., and Bontcheva, K. (2013).
\newblock Twitter part-of-speech tagging for all: Overcoming sparse and noisy
  data.
\newblock In {\em Proceedings of the International Conference on Recent
  Advances in Natural Language Processing}, Hissar, Bulgaria. Association for
  Computational Linguistics.

\bibitem[Desai and Narvekar, 2015]{desai2015normalization}
Desai, N. and Narvekar, M. (2015).
\newblock Normalization of noisy text data.
\newblock {\em Procedia Computer Science}, 45:127--132.

\bibitem[Durrani and Hussain, 2010]{durrani-hussain:2010:NAACLHLT}
Durrani, N. and Hussain, S. (2010).
\newblock Urdu word segmentation.
\newblock In {\em Proceedings of the Annual Conference of the North American
  Chapter of the Association for Computational Linguistics -- Human Language
  Technologies}, pages 528--536, Los Angeles, California, USA. Association for
  Computational Linguistics.

\bibitem[Durrani et~al., 2010]{durrani-EtAl:2010:ACL}
Durrani, N., Sajjad, H., Fraser, A., and Schmid, H. (2010).
\newblock Hindi-to-{U}rdu machine translation through transliteration.
\newblock In {\em Proceedings of the 48th Annual Meeting of the Association for
  Computational Linguistics}, pages 465--474, Uppsala, Sweden. Association for
  Computational Linguistics.

\bibitem[Fossati and Di~Eugenio, 2007]{fossati2007mixed}
Fossati, D. and Di~Eugenio, B. (2007).
\newblock A mixed trigrams approach for context sensitive spell checking.
\newblock In {\em Computational Linguistics and Intelligent Text Processing},
  pages 623--633. Springer.

\bibitem[Gouws et~al., 2011]{gouws2011unsupervised}
Gouws, S., Hovy, D., and Metzler, D. (2011).
\newblock Unsupervised mining of lexical variants from noisy text.
\newblock In {\em Proceedings of the 1st workshop on Unsupervised Learning in
  Natural Language Processing}, pages 82--90, Edinburgh, Scotland. Association
  for Computational Linguistics.

\bibitem[Hall and Dowling, 1980]{DBLP:journals/csur/HallD80}
Hall, P. A.~V. and Dowling, G.~R. (1980).
\newblock Approximate string matching.
\newblock {\em Association for Computing Machinery Computing Surveys},
  12(4):381--402.

\bibitem[Han et~al., 2012]{han2012automatically}
Han, B., Cook, P., and Baldwin, T. (2012).
\newblock Automatically constructing a normalisation dictionary for microblogs.
\newblock In {\em Proceedings of the Joint Conference on Empirical Methods in
  Natural Language Processing and Computational Natural Language Learning},
  pages 421--432, Jeju Island, Korea. Association for Computational
  Linguistics.

\bibitem[Han et~al., 2013]{han2013lexical}
Han, B., Cook, P., and Baldwin, T. (2013).
\newblock Lexical normalization for social media text.
\newblock {\em Association for Computing Machinery Transactions on Intelligent
  Systems and Technology}, 4(1):5.

\bibitem[Han, 2005]{dm_book}
Han, J. (2005).
\newblock {\em Data Mining: Concepts and Techniques}.
\newblock Morgan Kaufmann Publishers Inc., San Francisco, California.

\bibitem[Hany~Hassan, 2013]{Hassan-socialtext}
Hany~Hassan, Arul~Menezes, H. H.~A. (2013).
\newblock Social text normalization using contextual graph random walks.
\newblock In {\em Proceedings of the 51st Annual Meeting of the Association for
  Computational Linguistics}, Sofia, Bulgaria.

\bibitem[Hassan et~al., 2009]{rank_weight}
Hassan, M.~T., Junejo, K.~N., and Karim, A. (2009).
\newblock Learning and predicting key web navigation patterns using bayesian
  models.
\newblock In {\em Computational Science and Its Applications}, pages 877--887.
  Springer.

\bibitem[Irvine et~al., 2012]{Irvine2012}
Irvine, A., Weese, J., and Callison-Burch, C. (2012).
\newblock Processing informal, {R}omanized {P}akistani text messages.
\newblock In {\em Proceedings of the 2nd Workshop on Language in Social Media},
  LSM '12, pages 75--78, Montreal, Canada. Association for Computational
  Linguistics.

\bibitem[Jiampojamarn et~al., 2007]{jiampojamarn2007}
Jiampojamarn, S., Kondrak, G., and Sherif, T. (2007).
\newblock Applying many-to-many alignments and hidden markov models to
  letter-to-phoneme conversion.
\newblock In {\em Proceedings of the Conference of the North American Chapter
  of the Association for Computational Linguistics -- Human Language
  Technologies: Main Conference}, pages 372--379, Rochester, New York, USA.
  Association for Computational Linguistics.

\bibitem[Jin, 2015]{jin:2015:WNUT}
Jin, N. (2015).
\newblock Ncsu-sas-ning: Candidate generation and feature engineering for
  supervised lexical normalization.
\newblock In {\em Proceedings of the Workshop on Noisy User-generated Text},
  pages 87--92, Beijing, China. Association for Computational Linguistics.

\bibitem[Kaufmann and Kalita, 2010]{kaufmann2010syntactic}
Kaufmann, M. and Kalita, J. (2010).
\newblock Syntactic normalization of twitter messages.
\newblock In {\em Proceedings of the International Conference on Natural
  Language Processing}, Kharagpur, India.

\bibitem[Khan and Karim, 2012]{DBLP:conf/ictai/KhanK12}
Khan, O. and Karim, A. (2012).
\newblock A rule-based model for normalization of {SMS} text.
\newblock In {\em Proceedings of Institute of Electrical and Electronics
  Engineers 24th International Conference on Tools with Artificial
  Intelligence}, pages 634--641, Athens, Greece.

\bibitem[Knuth, 1973]{soundexbook}
Knuth, D.~E. (1973).
\newblock {\em The Art of Computer Programming: Volume 3, Sorting and
  Searching}.
\newblock Addison-Wesley.

\bibitem[Lafferty, 2001]{Lafferty01conditionalrandom}
Lafferty, J. (2001).
\newblock Conditional random fields: Probabilistic models for segmenting and
  labeling sequence data.
\newblock In {\em Proceedings of the 18th International Conference on Machine
  Learning}, pages 282--289, San Francisco, California, USA.

\bibitem[Levandowsky and Winter, 1971]{levandowsky1971}
Levandowsky, M. and Winter, D. (1971).
\newblock Distance between sets.
\newblock {\em Nature}, 234:34--35.

\bibitem[Lewis, 2009]{ethnologue}
Lewis, M.~P., editor (2009).
\newblock {\em Ethnologue: Languages of the World}.
\newblock SIL International, Dallas, Texas, USA, {Sixteenth} edition.

\bibitem[Li and Liu, 2014]{li-liu:2014:P14-3}
Li, C. and Liu, Y. (2014).
\newblock Improving text normalization via unsupervised model and
  discriminative reranking.
\newblock In {\em Proceedings of the Association for Computational Linguistics:
  Student Research Workshop}, pages 86--93, Baltimore, Maryland, USA.
  Association for Computational Linguistics.

\bibitem[Ling et~al., 2013]{ling2013paraphrasing}
Ling, W., Dyer, C., Black, A.~W., and Trancoso, I. (2013).
\newblock Paraphrasing 4 microblog normalization.
\newblock In {\em Proceedings of the Conference on Empirical Methods in Natural
  Language Processing}, pages 73--84, Seattle, Washington, USA.

\bibitem[Lusetti et~al., 2018]{Massimo18}
Lusetti, M., Ruzsics, T., G{\"o}hring, A., Samard{\v{z}}i{\'{c}}, T., and
  Stark, E. (2018).
\newblock Encoder-decoder methods for text normalization.
\newblock In {\em Proceedings of the 5th Workshop on Natural Language
  Processing for Similar Languages, Varieties and Dialects}, pages 18--28.
  Association for Computational Linguistics.

\bibitem[Mikolov et~al., 2013]{DBLP:journals/corr/MikolovSCCD13}
Mikolov, T., Sutskever, I., Chen, K., Corrado, G., and Dean, J. (2013).
\newblock Distributed representations of words and phrases and their
  compositionality.
\newblock In {\em Proceedings of the 26th International Conference on Neural
  Information Processing Systems}, page 3111–3119, Red Hook, New York, USA.
  Curran Associates Inc.

\bibitem[Naseem, 2004]{Tahira2004}
Naseem, T. (2004).
\newblock A hybrid approach for {Urdu} spell checking.
\newblock Master's thesis, Master of Science (Computer Science) thesis at the
  National University of Computer \& Emerging Sciences.

\bibitem[Naseem and Hussain, 2007]{journals/lre/NaseemH07}
Naseem, T. and Hussain, S. (2007).
\newblock A novel approach for ranking spelling error corrections for {U}rdu.
\newblock {\em Language Resources and Evaluation}, 41(2):117--128.

\bibitem[Nelder and Mead, 1965]{NelderMead65}
Nelder, J.~A. and Mead, R. (1965).
\newblock A simplex method for function minimization.
\newblock {\em Computer Journal}, 7:308--313.

\bibitem[Och and Ney, 2003]{och2003systematic}
Och, F.~J. and Ney, H. (2003).
\newblock A systematic comparison of various statistical alignment models.
\newblock {\em Computational Linguistics}, 29(1):19--51.

\bibitem[Paltoglou and Thelwall, 2012]{paltoglou2012twitter}
Paltoglou, G. and Thelwall, M. (2012).
\newblock {Twitter, MySpace, Digg}: Unsupervised sentiment analysis in social
  media.
\newblock {\em Association for Computing Machinery Transactions on Intelligent
  Systems and Technology}, 3(4):66.

\bibitem[Pennell and Liu, 2011]{pennell2011character}
Pennell, D. and Liu, Y. (2011).
\newblock A character level machine translation approach for normalization of
  {SMS} abbreviations.
\newblock In {\em Proceedings of the 5th International Joint Conference on
  Natural Language Processing}, pages 974--982, Chiang Mai, Thailand.

\bibitem[Philips, 1990]{clm/philips90}
Philips, L. (1990).
\newblock Hanging on the metaphone.
\newblock {\em Computer Language Magazine}, 7(12):39--44.

\bibitem[Pinto et~al., 2012]{DBLPconftsdPintoAAGLJ12}
Pinto, D., Ayala, D.~V., Alem{\'{a}}n, Y., G{\'{o}}mez{-}Adorno, H., Loya, N.,
  and Jim{\'{e}}nez{-}Salazar, H. (2012).
\newblock The soundex phonetic algorithm revisited for {SMS} text
  representation.
\newblock In {\em Proceedings of the 15th International Conference on Text,
  Speech and Dialogue}, pages 47--55, Brno, Czech Republic.

\bibitem[Prochasson et~al., 2007]{prochasson07language}
Prochasson, E., Viard{-}Gaudin, C., and Morin, E. (2007).
\newblock Language models for handwritten short message services.
\newblock In {\em Proceedings of the 9th International Conference on Document
  Analysis and Recognition}, pages 83--87, Parana, Brazil. Institute of
  Electrical and Electronics Engineers Computer Society.

\bibitem[Rafae et~al., 2015]{rafaeQUKSK15}
Rafae, A., Qayyum, A., Uddin, M.~M., Karim, A., Sajjad, H., and Kamiran, F.
  (2015).
\newblock An unsupervised method for discovering lexical variations in {R}oman
  {U}rdu informal text.
\newblock In {\em Proceedings of the Conference on Empirical Methods in Natural
  Language Processing}, pages 823--828, Lisbon, Portugal. Association for
  Computational Linguistics.

\bibitem[Rangarajan~Sridhar, 2015]{sridhar2015unsupervised}
Rangarajan~Sridhar, V.~K. (2015).
\newblock Unsupervised text normalization using distributed representations of
  words and phrases.
\newblock In {\em Proceedings of the 1st Workshop on Vector Space Modeling for
  Natural Language Processing}, pages 8--16, Denver, Colorado, USA. Association
  for Computational Linguistics.

\bibitem[Rangarajan~Sridhar et~al., 2014]{Sridhar2014}
Rangarajan~Sridhar, V.~K., Chen, J., Bangalore, S., and Shacham, R. (2014).
\newblock A framework for translating {SMS} messages.
\newblock In {\em Proceedings of the 25th International Conference on
  Computational Linguistics: Technical Papers}.

\bibitem[Ristad and Yianilos, 1998]{RYsed98}
Ristad, E.~S. and Yianilos, P.~N. (1998).
\newblock Learning string edit distance.
\newblock {\em Institute of Electrical and Electronics Engineers Transactions
  on Pattern Recognition and Machine Intelligence}, 20(5):522--532.

\bibitem[Roy et~al., 2013]{roylexicon}
Roy, S., Dhar, S., Bhattacharjee, S., and Das, A. (2013).
\newblock A lexicon based algorithm for noisy text normalization as
  pre-processing for sentiment analysis.
\newblock {\em International Journal of Research in Engineering and
  Technology}, 02.

\bibitem[Sajjad et~al., 2011]{sajjad:acl11}
Sajjad, H., Fraser, A., and Schmid, H. (2011).
\newblock An algorithm for unsupervised transliteration mining with an
  application to word alignment.
\newblock In {\em Proceedings of the 49th Conference of the Association for
  Computational Linguistics -- Human Language Technologies}, Portland, Oregon,
  USA.

\bibitem[Sajjad and Schmid, 2009]{SajjadSchmid:09}
Sajjad, H. and Schmid, H. (2009).
\newblock Tagging {U}rdu text with parts of speech: A tagger comparison.
\newblock In {\em Proceedings of the 12th Conference of the European Chapter of
  the Association for Computational Linguistics}, pages 692--700, Athens,
  Greece.

\bibitem[Sajjad et~al., 2017]{sajjad2017statistical}
Sajjad, H., Schmid, H., Fraser, A., and Sch{\"u}tze, H. (2017).
\newblock Statistical models for unsupervised, semi-supervised and supervised
  transliteration mining.
\newblock {\em Computational Linguistics}, 43(2).

\bibitem[Sidarenka et~al., 2013]{sidarenka2013rule}
Sidarenka, U., Scheffler, T., and Stede, M. (2013).
\newblock Rule-based normalization of {G}erman twitter messages.
\newblock In {\em Proceedings of the Gesellschaft fur Sprachtechnologie und
  Computerlinguistik Workshop Verarbeitung und Annotation von Sprachdaten aus
  Genres internetbasierter Kommunikation}, Darmstadt, Germany.

\bibitem[Singh et~al., 2018]{Singh18}
Singh, R., Choudhary, N., and Shrivastava, M. (2018).
\newblock Automatic normalization of word variations in code-mixed social media
  text.
\newblock {\em Computing Research Repository}, abs/1804.00804.

\bibitem[Sproat et~al., 2001]{Sproat99normalizationof}
Sproat, R., Black, A.~W., Chen, S.~F., Kumar, S., Ostendorf, M., and Richards,
  C. (2001).
\newblock Normalization of non-standard words.
\newblock {\em Computer Speech {\&} Language}, 15(3):287--333.

\bibitem[Sproat and Jaitly, 2017]{Sproat17}
Sproat, R. and Jaitly, N. (2017).
\newblock An rnn model of text normalization.
\newblock In {\em Proceedings of Interspeech}, Stockholm, Sweden.

\bibitem[Taft, 1970]{NYSIIS}
Taft, R. (1970).
\newblock Name search techniques.
\newblock {\em Special report. Bureau of Systems Development, New York State
  Identification and Intelligence System}.

\bibitem[Vilain et~al., 1995]{M95-1005}
Vilain, M., Burger, J., Aberdeen, J., Connolly, D., and Hirschman, L. (1995).
\newblock A model-theoretic coreference scoring scheme.
\newblock In {\em Proceedings of the 6th Message Understanding Conference},
  pages 45--52, Columbia, Maryland, USA.

\bibitem[Wang, 2009]{DBLP:reference/dataware/2009}
Wang, J., editor (2009).
\newblock {\em Encyclopedia of Data Warehousing and Mining, Second Edition {(4}
  Volumes)}.
\newblock {IGI} Global.

\bibitem[Wang and Ng, 2013]{wang-ng:2013:NAACL-HLT}
Wang, P. and Ng, H.~T. (2013).
\newblock A beam-search decoder for normalization of social media text with
  application to machine translation.
\newblock In {\em Proceedings of the Conference of the North American Chapter
  of the Association for Computational Linguistics -- Human Language
  Technologies}, pages 471--481, Atlanta, Georgia, USA. Association for
  Computational Linguistics.

\bibitem[Wei et~al., 2011]{wei2011exploring}
Wei, Z., Zhou, L., Li, B., Wong, K.-F., Gao, W., and Wong, K.-F. (2011).
\newblock Exploring tweets normalization and query time sensitivity for twitter
  search.
\newblock In {\em Proceedings of the 20th Text Retrieval Conference},
  Gaithersburg, Maryland, USA.

\bibitem[Yang and Eisenstein, 2013]{Yang-alog-linear}
Yang, Y. and Eisenstein, J. (2013).
\newblock A log-linear model for unsupervised text normalization.
\newblock In {\em Proceedings of the Conference on Empirical Methods in Natural
  Language Processing: Meeting of SIGDAT, a Special Interest Group of the
  Association for Computational Linguistics}, pages 61--72, Seattle,
  Washington, USA.

\bibitem[Zhang et~al., 2015]{zhang2015normalization}
Zhang, X., Song, J., He, Y., and Fu, G. (2015).
\newblock Normalization of homophonic words in {C}hinese microblogs.
\newblock In {\em Intelligent Computation in Big Data Era}, pages 177--187.
  Springer.

\end{thebibliography}

\end{document}